\documentclass[lettersize,journal]{IEEEtran}
\usepackage{amsmath,amsfonts}
\usepackage{algorithmic}
\usepackage{algorithm}
\usepackage{array}
\usepackage[caption=false,font=normalsize,labelfont=sf,textfont=sf]{subfig}
\usepackage{textcomp}
\usepackage{booktabs}
\usepackage{stfloats}
\usepackage{url}
\usepackage{verbatim}
\usepackage{graphicx}
\usepackage{cite}
\usepackage{multirow}
\usepackage{booktabs}
\begin{document}

\title{ReasonCD: A Multimodal Reasoning Large Model for Implicit Change-of-Interest Semantic Mining}

\author{Zhenyang Huang, Xiao Yu, Yi Zhang, Decheng Wang, Hang Ruan 
        % <-this % stops a space
% 致谢
\thanks{The work was supported in part by xxxxx, in part by the xxxx. (Corresponding author: Hang Ruan.)}% <-this % stops a space
\thanks{Zhenyang Huang, Xiao Yu, Yi Zhang, Decheng Wang and Hang Ruan are with the Beijing Institute of Tracking and Telecommunications Technology.(e-mail: 704835342@qq.com; yuxiao801026@126.com; MinnieZ423@163.com; 18810836867@163.com; ruanhang\_bds@163.com;).}
}

% The paper headers
\markboth{Journal of \LaTeX\ Class Files,~Vol.~14, No.~8, August~2021}%
{Shell \MakeLowercase{\textit{et al.}}: A Sample Article Using IEEEtran.cls for IEEE Journals}

% \IEEEpubid{0000--0000/00\$00.00~\copyright~2021 IEEE}
% Remember, if you use this you must call \IEEEpubidadjcol in the second
% column for its text to clear the IEEEpubid mark.

\maketitle

\begin{abstract}
Remote sensing image change detection is one of the fundamental tasks in remote sensing intelligent interpretation. Its core objective is to identify changes within change regions of interest (CRoI). Current multimodal large models encode rich human semantic knowledge, which is utilized for guidance in tasks such as remote sensing change detection. However, existing methods that use semantic guidance for detecting users' CRoI overly rely on explicit textual descriptions of CRoI, leading to the problem of near-complete performance failure when presented with implicit CRoI textual descriptions. This paper proposes a multimodal reasoning change detection model named ReasonCD, capable of mining users' implicit task intent. The model leverages the powerful reasoning capabilities of pre-trained large language models to mine users' implicit task intents and subsequently obtains different change detection results based on these intents. Experiments on public datasets demonstrate that the model achieves excellent change detection performance, with an F1 score of 92.1\% on the BCDD dataset. Furthermore, to validate its superior reasoning functionality, this paper annotates a subset of reasoning data based on the SECOND dataset. Experimental results show that the model not only excels at basic reasoning-based change detection tasks but can also explain the reasoning process to aid human decision-making.

%a Siamese foreground association-driven hard case sample optimization network (HASNet)

\end{abstract}


\section{Introduction}
\IEEEPARstart{R}{emote} sensing image change detection is a significant theoretical problem in fields such as remote sensing image understanding and computer vision. It is a key technology supporting major national needs including national defense security\cite{guofanganquan}, disaster emergency response\cite{zaihai}, and land resource surveys\cite{ziyuan1,ziyuan2}.The advent of CLIP\cite{clip} in 2021 and ChatGPT\cite{chatgpt} in 2022 fundamentally altered the technical roadmap for remote sensing change detection. 

To date, researchers have proposed various methods for change detection tasks. Zhao et al. categorized current change detection methods into vision-first(Vifi-CD) and semantics-first(Sefi-CD) paradigms\cite{seficd}. Addressing the issue in the visual-first paradigm of being unable to flexibly detect different CRoI changes according to varying user task requirements, they proposed a semantics-first paradigm method named AUWCD\cite{seficd}. This method consists of three modules: a Semantic Align module, a CRoI Segment module, and a change detection post-processing module. It can dynamically perceive semantic information of changes of interest and obtain corresponding change detection results based on simple textual descriptions of different user CRoIs.However, the effectiveness of existing semantics-first paradigm methods depends on the precision and richness of the CRoI textual description; more specific and richer descriptions lead to better change detection performance. Consequently, current SeFi-CD methods are only suitable for change detection tasks where users provide explicit CRoI instructions. When the given CRoI textual description is overly implicit, their change detection performance degrades significantly. For example, before performing a change detection task, if it is known that a region experienced heavy rainfall, and the model's CRoI textual instruction is "changes caused by heavy rainfall," the performance of current SeFi-CD methods would drop substantially or even fail completely. This is because they perform well only with concrete CRoI descriptions like "lake" but lack the ability to reason about such implicit CRoIs.

\begin{figure}[H]%[h]  Figure 1
%  \centering
    \includegraphics[width=0.5\textwidth]{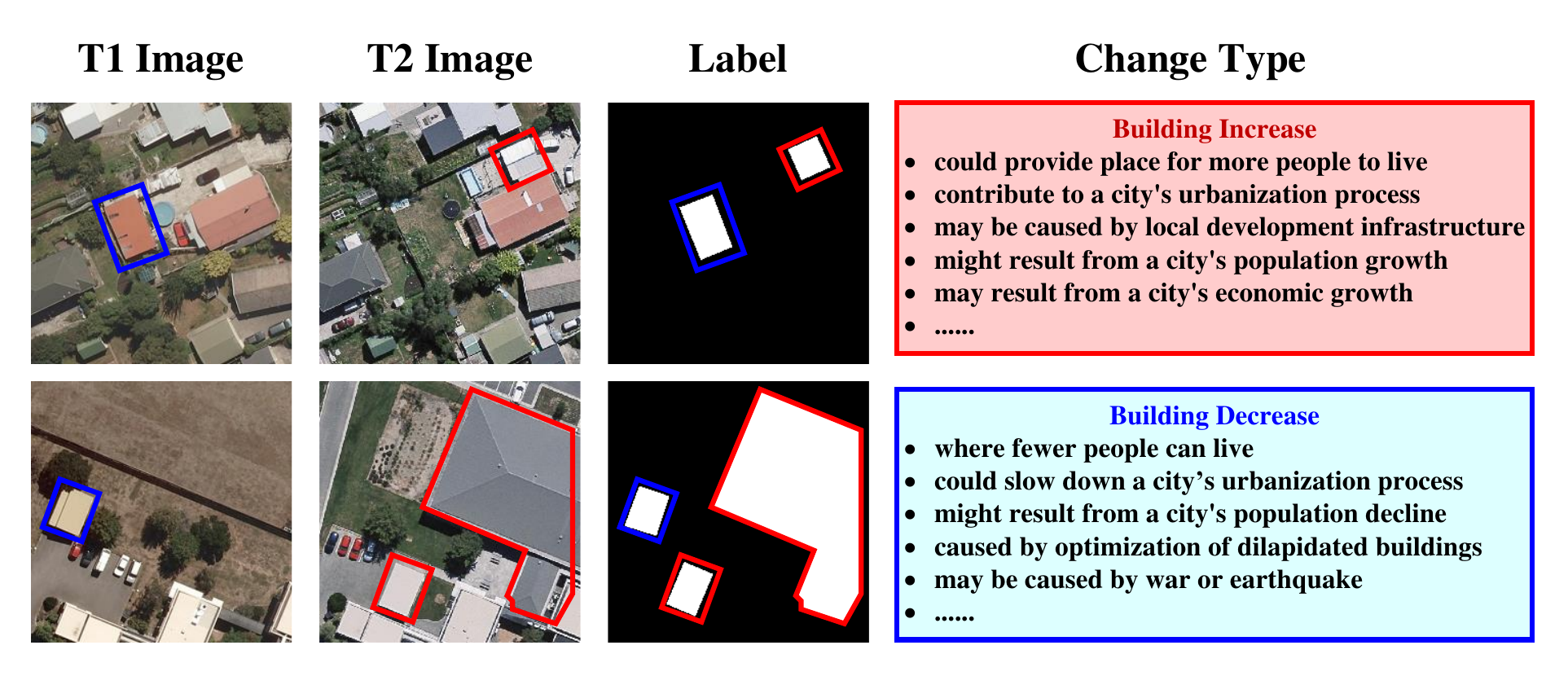}
    \caption{Explicit and Implicit Semantic Expressions of CRoI, Using Buildings as an Example.}
    \label{fig1}
\end{figure}

Nevertheless, many change detection task requirements are implicit. To clarify this concept, this paper categorizes the semantic ambiguity in current semantics-first paradigm change detection tasks into explicit semantic understanding and implicit semantic understanding. Therefore, AUWCD\cite{seficd} naturally falls under the category of semantics-first change detection paradigm methods for explicit semantic understanding. As shown in Figure 1, for a building change detection task, the implicit ways to express "building" as the CRoI are diverse. For building changes alone, this includes two types of temporal changes: building increase or decrease. For building increase, descriptions could include "changes that provide more living areas for people," "changes that promote urbanization," "changes brought by local construction," "changes caused by local population growth," "changes caused by local economic growth," etc. For building decrease, descriptions could be "changes that would reduce people's living areas," "changes that slow down urbanization," "changes brought by optimization and renovation of old buildings," "changes possibly caused by war or natural disasters," etc. Therefore, proposing a change detection model with implicit CRoI reasoning capability is crucial for developing next-generation remote sensing intelligent perception systems. Such a model should possess at least the following capabilities:
\begin{itemize}
\item{Support multimodal input: at least supporting text and bitemporal image inputs.}
\item{Ability to infer a change map by combining images and implicit CRoI descriptions.}
\item{SeFi-CD paradigm: Still obtain different CRoI change detection results based on different CRoI textual descriptions without additional model training.}
\end{itemize}

Based on the described problem and motivation, the main contributions of this article are as follows:
\begin{enumerate}
  \item{\textbf{Definition of Implicit CRoI Semantic Reasoning Problem}: This paper categorizes the semantic ambiguity in current SeFi-CD paradigm change detection tasks as explicit semantics and implicit semantics. Addressing the issue where current SeFi-CD methods can only perform change detection for explicit semantics and nearly fail for implicit semantics, it defines the change detection task of implicit CRoI semantic reasoning.}
  \item{\textbf{ReasonCD Network}: This paper proposes the first multimodal change detection large model under the SeFi-CD paradigm for implicit CRoI reasoning—ReasonCD. ReasonCD uses a pre-trained large language model as the core reasoning component, supports multimodal input of text and bitemporal images, employs text to model the implicit CRoI requiring reasoning, and uses the LLM to infer high-dimensional features representing the implicit semantics. Finally, a ChangeMap Decoder with a Multi-scale Feature Fusion Module is designed to decode high-quality change maps based on the inferred high-dimensional implicit semantic features.}
  \item{\textbf{Validation of ReasonCD's Effectiveness and Feasibility}: Experiments on public datasets show that ReasonCD not only achieves good results in simple single-CRoI fitting tasks, with an F1 score reaching 0.921 on the BCDD dataset, surpassing many classical change detection methods, but more importantly, ReasonCD can simultaneously achieve reasoning for both explicit and implicit CRoIs, obtaining corresponding change maps based on different implicit CRoI textual descriptions. It can also explain the reasoning process as needed.}
\end{enumerate}

The rest of this paper is organized as follows. Section 2 reviews related work. Section 3 describes our method. Section 4 describes the design of the validation experiments, followed by the analysis and discussion. Section 5 concludes the paper and proposes research directions for future work.

\section{Related works}
As this paper primarily investigates issues with current SeFi-CD paradigm methods, this section first reviews existing SeFi-CD paradigm change detection methods. Secondly, since the core of ReasonCD's reasoning is a pre-trained large language model, this section also reviews related work on multimodal reasoning large models based on pre-trained LLMs. Details are as follows.

\subsection{Semantics-First Change Detection Methods}
The SeFi-CD paradigm was first proposed by Zhao et al., who correspondingly introduced a method under this paradigm named AUWCD\cite{seficd}. This method consists of three modules: a Semantic Align module, a CRoI Segment module, and a change detection post-processing module. It can dynamically perceive semantic information of changes of interest and obtain corresponding change detection results based on simple textual descriptions of different user CRoIs. The change-agent\cite{change_agent} proposed by Liu et al. also essentially belongs to this paradigm. Trained on datasets annotated with roads and buildings, it can interactively perform change detection for roads, buildings, or both based on user requirements. Wang et al. proposed the Change Knowledge-guided Vision-Language Remote Sensing Change Detection method (CKCD), which introduces change knowledge as linguistic information to address limitations of purely visual methods\cite{ckcd}. Zhu et al. proposed Semantic-CD\cite{semanticcd}, a method also belonging to the SeFi-CD paradigm. Similar to AUWCD, it leverages CLIP's broad vocabulary knowledge to enhance cross-category general understanding, enabling finer-grained semantic change detection tasks.

Current SeFi-CD paradigm change detection methods typically combine simple textual descriptions with CLIP's text encoder, capable only of change detection tasks with explicit semantic understanding as described above, or improving change detection performance by introducing semantics. However, as noted in Zhao et al.'s work\cite{seficd}, this approach of combining simple text descriptions with CLIP demands high accuracy in the textual description and struggles to understand implicit CRoI semantics for some complex user task requirements. Therefore, we need to propose a method to address the issue that current SeFi-CD change detection methods find difficult: performing change detection involving implicit CRoI semantic reasoning.

\subsection{Multimodal Reasoning Large Models}
Leveraging the powerful reasoning capabilities of large language models, researchers subsequently explored methods to transfer this reasoning ability to the visual domain, developing various multimodal reasoning large models based on LLMs. Flamingo, proposed by Alayrac et al., uses a Cross-Attention structure to introduce visual attention, enabling in-context learning for vision and text\cite{flamingo}. Models like BLIP-2\cite{blip2} by Li et al. and mPLUG-OWL\cite{mplug} by Ye et al. designed a Q-Former or similar module to feed the most useful visual feature embeddings into the large language model to obtain the desired textual output. Otter, proposed by Li et al., uses their proposed MIMIC-IT dataset for in-context instruction tuning, further enhancing the model's robustness for few-shot transfer\cite{otter}. Subsequently, LLaVA\cite{llava}by Liu et al. and MiniGPT-4\cite{minigpt}by Zhu et al. first perform image-text feature alignment and then use instruction tuning for downstream tasks. Furthermore, many works utilize prompt engineering to connect independent modules via API calls, fully leveraging each module's ability to handle data from its respective modality\cite{bs85,bs86,bs87,bs88,bs89}, though these methods often cannot be trained end-to-end.

Later, some research within the domain focused on the intersection of multimodal large language models and vision tasks. VisionLLM, proposed by Wang et al., provides a flexible vision interface for multiple vision-centric tasks through instruction tuning but cannot fully utilize the LLM for reasoning in complex tasks\cite{visionllm}. Kosmos-2, proposed by Peng et al., constructs a large-scale image-text dataset with bounding boxes, injecting object localization reasoning capability into the multimodal large language model\cite{kosmos}. DetGPT, proposed by Pi et al., bridges a fixed multimodal large language model with an open-vocabulary detector, enabling object detection based on user instructions\cite{detgpt}. Later, LISA, proposed by Lai et al., implements reasoning segmentation, capable of inferring masks for user-specified targets based on need, deepening the model's understanding of images and text\cite{lisa}. Inspired by LISA, Wei et al. applied the same idea to reasoning object detection, proposing Lenna, a multimodal reasoning object detection large model that operates based on user instructions\cite{lenna}.

Therefore, outside the change detection field, many researchers have investigated the reasoning capabilities of multimodal large models. However, within the change detection field, there has been no research on multimodal reasoning large models for change detection.

In summary, existing work struggles to achieve the aforementioned implicit CRoI semantic reasoning change detection task, and there is indeed a lack of research on multimodal reasoning large models in the change detection domain. Therefore, we need to propose a method based on multimodal data and reasoning large models to accomplish the implicit CRoI semantic reasoning change detection task.

\section{Methodology}

\begin{figure*}[ht]%[h]  Figure 2
\centering
\includegraphics[width=7in]{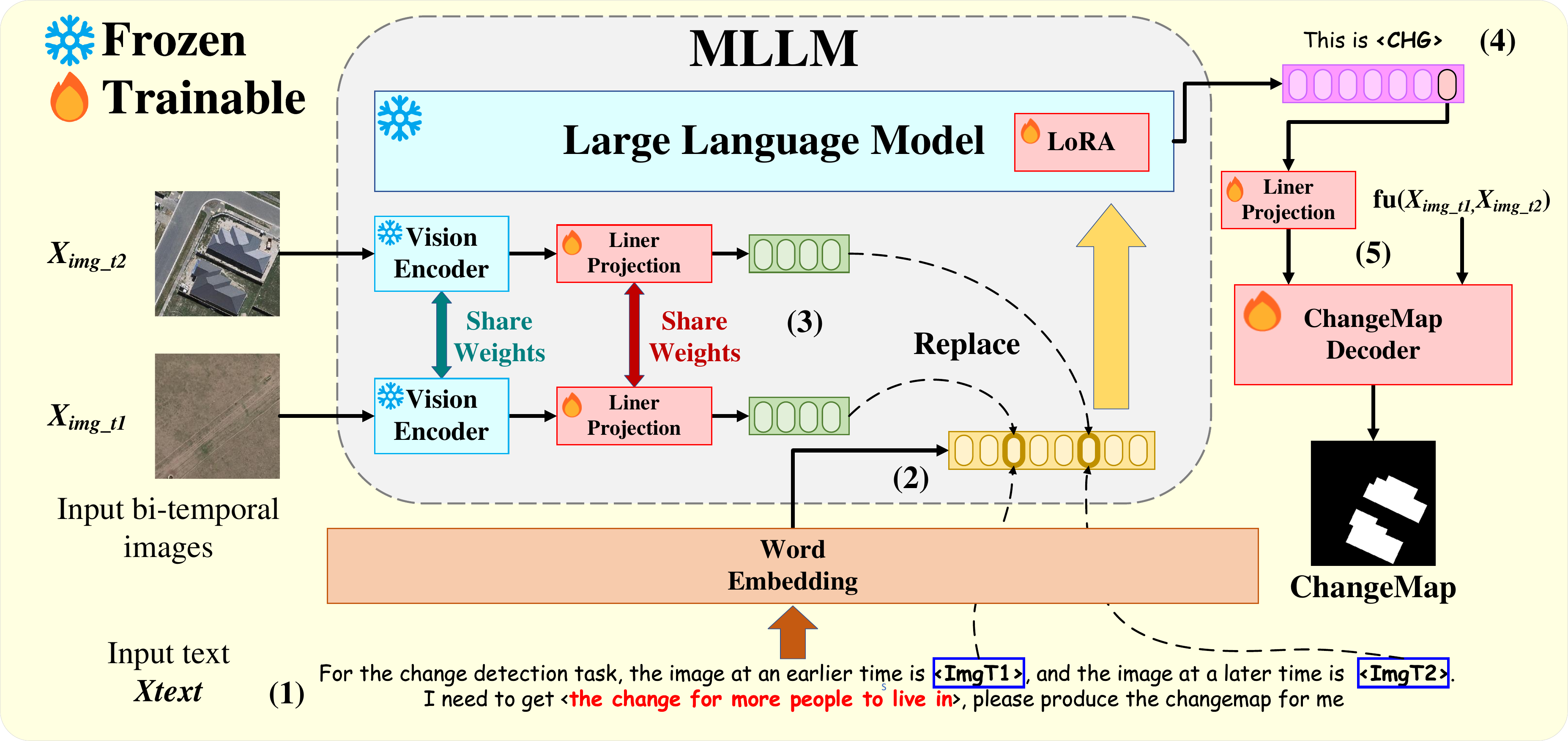}
\caption{Schematic diagram of the overall ReasonCD structure. ReasonCD is mainly divided into five parts: (1) Change detection task text modeling; (2) Text tokenization and embedding; (3) Bitemporal image embedding; (4) Multimodal token reasoning; (5) Change map decoding based on the $<CHG>$ token.}
\label{fig2}
\end{figure*}

\subsection{Problem Definition}
We define the implicit CRoI semantic reasoning change detection as a process with multimodal input and multimodal output, which can be expressed as:

\begin{equation}
\label{eq1}
Y_{\text{text}}, Y_{\text{$changemap$}}=f\left(X_{\text{$img_{t1}$}}, X_{\text{$img_{t2}$}}, X_{\text{text }}\right)
\end{equation}

Here, $X_{img_{t1}}$, $X_{img_{t2}}$, $X_{text}$ are multimodal inputs, representing the input image at time T1, the input image at time T2, and the input CRoI textual description (supporting both explicit and implicit descriptions), respectively; $f$ represents the reasoning-based change detection model or algorithm; $Y_{text}$, $Y_{changemap}$ are the multimodal outputs after reasoning, representing the textual output and the corresponding CRoI change map, respectively.

\subsection{The ReasonCD Framework}
The overall structure of ReasonCD can be divided into five parts, as shown in Figure \ref{fig2}:
\subsubsection{\textbf{Change Detection Task Text Modeling}}
This part addresses the lack of temporal relationship and task information in the text modality of SeFi-CD by proposing a change detection text modeling strategy: using $<ImgT1>$ and $<ImgT2>$ tokens to embed image temporal order, explicitly stated in the template; combining task description and result prompts to enhance the LLM's task awareness; and using $<CRoI>$ tokens to highlight the region description requiring reasoning, thereby improving text embedding quality and model reasoning effectiveness. Note that the embeddings of the $<ImgT1>$ and $<ImgT2>$ tokens do not contain image information at this stage; they only serve as placeholders.
\subsubsection{\textbf{Text Tokenization and Embedding}}
This part primarily tokenizes the modeled change detection task text and then uses a text tokenizer to obtain text embeddings. Additionally, we reintroduce 8 special token markers to enhance the model's structured understanding of the task text.
\subsubsection{\textbf{Bitemporal Image Embedding}}
This part primarily uses a parameter-frozen Vision Encoder to extract features from the bitemporal images. Then, a trainable Linear Projection maps the extracted visual features to the same dimension as the text embedding, obtaining Bi-temporal image embeddings. Since the embeddings of the special tokens $<ImgT1>$ and $<ImgT2>$ lack image information, we replace these token embeddings with the Bi-temporal image embeddings. This ensures the final task text modeling embedding contains multimodal visual and textual information, aligning with the current mainstream modality alignment approach.
\subsubsection{\textbf{Multimodal Token Reasoning}}
The primary function of this part is to reason over the constructed multimodal tokens and mine implicit CRoI semantics. Many works have demonstrated the reasoning capabilities of pre-trained LLMs.  Therefore, we use a pre-trained LLM—\textbf{LLaMA2}\cite{llama2}—as the core reasoning component in this part. During training, to reduce GPU memory usage, we employ LoRA (Low-Rank Adaptation)\cite{lora}for efficient fine-tuning. After fine-tuning, the output text from LLaMA2 will contain a $<CHG>$ token, which represents the inferred implicit CRoI semantics. This paper posits that the high-dimensional token embedding of this marker represents sufficiently rich implicit CRoI semantic information. To avoid information loss from repeated conversions between high-dimensional features → low-dimensional text → high-dimensional features\cite{rsprompter}, we extract the last-layer Embedding feature corresponding to the $<CHG>$ token as the high-level semantic feature representing the reasoned CRoI.
\subsubsection{\textbf{Change Map Decoding Based on the $<CHG>$ Token}}
The main function of this part is to decode the final Change Map corresponding to the implicit CRoI textual description, using the aforementioned last-layer Embedding of the $<CHG>$ token and the visual features of the bitemporal images via a ChangeMap Decoder. Simultaneously, we propose a novel \textbf{Multi-scale Feature Fusion Module $fu(\bullet)$}, which integrates high-level features extracted by a deep Vision Transformer\cite{vit}and multi-level low-level visual features extracted by ResNet-FPN\cite{fpn}, to facilitate ChangeMap training convergence and decode higher-quality Change Maps. 

To provide a clearer introduction to the computational flow of the ReasonCD algorithm, the following sections will focus on key internal processes and computational details.

\subsection{Change Detection Task Text Modeling}
The design motivation for this part primarily considers two common issues in current SeFi-CD methods when using the text modality:
\begin{itemize}
\item{Lack of temporal relationship between bitemporal images.}
\item{Lack of more specific change detection task information.}
\end{itemize}
The specific method for change detection task text modeling designed in ReasonCD to address these two issues is as follows:

\textbf{Temporal Relationship Between Bitemporal Images}. From Figure \ref{fig1}, it can be seen that the temporal relationship of bitemporal images is crucial in implicit CRoI semantic reasoning change detection tasks, directly determining the detection results (e.g., building increase or decrease). Therefore, to model this temporal relationship, this paper uses two special tokens, $<ImgT1>$ and $<ImgT2>$, to represent the positions of the bitemporal images in the text (i.e., as placeholders). As shown in Figure \ref{fig2}, the prompt template provides additional explanation before these special tokens, indicating that the earlier image is $<ImgT1>$ and the later image is $<ImgT2>$. Although these special tokens are simple words without specific image information, their design purpose is to encode this contextual temporal information into the text embeddings during subsequent text embedding, thereby promoting model training convergence.

\textbf{More Specific Change Detection Task Information}. Since ReasonCD subsequently uses a pre-trained large language model as the reasoning core, and pre-trained LLMs, trained on vast corpora, have clear definitions and concepts for change detection tasks, task information is incorporated by adding a description of the task at the beginning of the prompt template, e.g., "For the Change Detection task, ...". Furthermore, to better inform the model of the required results for the change detection task, we add the task requirement at the end of the prompt template, e.g., "Please produce the changemap for me."

Additionally, to strengthen the model's attention to the implicit CRoI textual description requiring reasoning, this paper adds two special tokens, $<CRoI>$ and $</CRoI>$, before and after the implicit CRoI textual description, to indicate the start and end positions of the description to be reasoned.

This section aims to address the two aforementioned common problems in current SeFi-CD methods through change detection text task modeling, obtaining higher-quality feature vectors during subsequent text embedding, and ultimately enabling ReasonCD to achieve better reasoning and change detection performance.

\subsection{Text Tokenization and Embedding}
After modeling the task text using the implicit CRoI textual description, a tokenizer is needed to segment the input text into tokens and encode these tokens to obtain corresponding text embeddings. The tokenizer in ReasonCD uses SentencePiece\cite{sentencepiece} to implement Byte Pair Encoding (BPE)\cite{bpe} for corpus tokenization.

SentencePiece is an unsupervised text tokenization tool capable of processing raw text directly. It learns the most common subword units from large-scale text data through statistical modeling, constructing a stable and compact vocabulary. BPE is an algorithm that iteratively merges the most frequent substrings. It initially splits text into individual characters and then iteratively merges the most frequently occurring adjacent symbols (or subwords) to build higher-level subword units. For words not present or with low frequency in the training data, BPE splits them into multiple common subword units, improving the model's generalization to rare or new words. During training, ReasonCD's tokenizer learns a fixed-size vocabulary via BPE, enabling the model to maintain sufficient expressive power while controlling vocabulary size and reducing computational burden. After training, the learned vocabulary size is 31,997. In addition, this paper adds 8 special tokens as shown in Table \ref{tab1}, making the final vocabulary size 32,005.

Tokens obtained by tokenizing the input text are typically IDs from the vocabulary. These IDs, usually represented as integers, are not suitable as high-dimensional vector inputs for the large language model to learn. Therefore, a separate embedding step is required, encoding these token IDs into corresponding feature vectors. Here, each token ID is mapped to a high-dimensional vector in a pre-trained Embedding Lookup Table (ELT) to obtain text embeddings.
% Please add the following required packages to your document preamble:
% \usepackage{booktabs}
% \usepackage{graphicx}
\begin{table}[]
\centering
\caption{Additional Special Tokens Table}
\label{tab1}
\resizebox{\columnwidth}{!}{%
\begin{tabular}{@{}ccc@{}}
\toprule
No.     & Special Token & Meaning
\\ \midrule
1 & $<s>$ & Input text start token  \\
2 & $</s>$ & Input text end token \\
3 & $<unk>$ & Unknown token marker \\
4 & $<ImgT1>$ & Placeholder token for T1 image \\
5 & $<ImgT2>$ & Placeholder token for T2 image \\
6 & $<CRoI>$ & Start token for implicit CRoI text description \\
7 & $</CRoI>$ & End token for implicit CRoI text description \\
8 & $<CHG>$ & Token for reasoned CRoI semantics \\
\bottomrule
\end{tabular}%
}
\end{table}

The ELT is a matrix of size (Vocabulary\_Size x Embedding\_Dimension), where each row represents the embedding vector of a word in the vocabulary. During model runtime, the IDs of a sentence's tokens are typically converted into one-hot encoding form and then multiplied by the ELT to obtain the final text embedding, expressed as:
% eq2
\begin{equation}
\label{eq2}
text\_token_{one-hot}=OneHot(text\_token_{ID})
\end{equation}
% eq3
\begin{equation}
\label{eq3}
{text\_token } _{embedding}=text\_token_{one-hot} \times ELT
\end{equation}
% text
where, $text\_token_{ID}\in\mathbb{R}^{num\_tokens\times1}$ represents the token IDs obtained after tokenizing the input text, and $num\_tokens$ is the number of tokens obtained after tokenization. $OneHot(\bullet)$ represents the transformation process from IDs to one-hot encoding. Since the total vocabulary size is 32,005, $text\_token_{one-hot}\in\mathbb{R}^{num\_tokens\times32005}$ represents the tokens in one-hot encoded form.$ELT\in\mathbb{R}^{32005\times4096}$, therefore, after matrix multiplication, $text\_token_{embedding}\in\mathbb{R}^{num\_tokens\times4096}$, represents the final text embedding.

\subsection{Bitemporal Image Embedding}
In the change detection task text modeling, only the special tokens $<ImgT1>$ and $<ImgT2>$ are used as placeholders in the input text, and $text\_token_{embedding}$ does not contain specific image features. Therefore, image embeddings need to be obtained and implanted into $text\_token_{embedding}$. Here, this paper selects CLIP\cite{clip} as the model for obtaining image embeddings.
\begin{figure}[H]%[h]  Figure 3
%  \centering
    \includegraphics[width=0.5\textwidth]{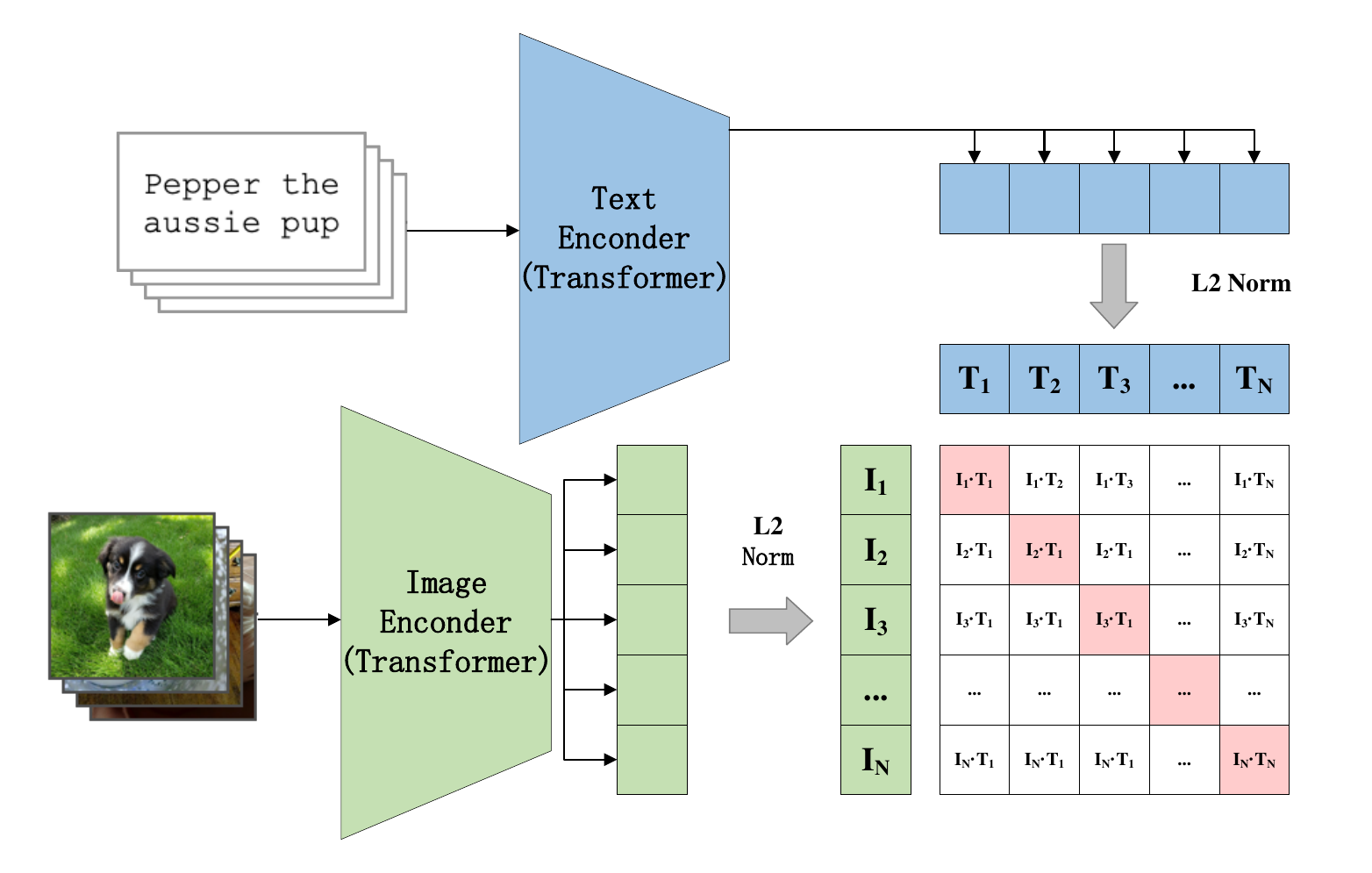}
    \caption{Schematic diagram of CLIP principle}
    \label{fig3}
\end{figure}
CLIP is a model released by OpenAI in early 2021 and is a classic work in the multimodal field in recent years. CLIP, trained on 400 million image-text pairs scraped from the internet, possesses powerful image-text understanding capabilities, making it very suitable for embedding its extracted visual features into $text\_token_{embedding}$. As shown in Figure \ref{fig3}, CLIP consists of encoders for both image and text modalities, both based on Transformer architecture. For a batch of N image-text pairs, images are encoded using CLIP's image encoder. After L2 normalization, the feature for each image is $I\in\mathbb{R}^{N\times dim\_I}$, where $dim\_I$ is the feature dimension after encoding. Similarly, CLIP's text encoder encodes the text descriptions of the images. After L2 normalization, the encoding for the text descriptions is $T\in\mathbb{R}^{N\times dim\_T}$, where $dim\_T$ is the feature dimension after encoding each description, equal to $dim\_I$. Then, matrix multiplication is used to obtain the similarity logits between each image and each text description in the batch:
% eq4
\begin{equation}
\label{eq4}
logits=I\times T^T\cdot e^t
\end{equation}
where t is the temperature controlling the parameter update intensity. Then, a contrastive learning approach can be employed to pull the similarity of paired image-text pairs closer and push the similarity of unpaired samples apart. Thus, the image features encoded by the trained CLIP image encoder can be aligned with text features and used for bitemporal image embedding. 

Since CLIP's image encoder is also based on the ViT\cite{vit} architecture, the visual feature used during its training is the last-layer embedding feature of the [CLS] token, losing much of the original image detail. Therefore, when extracting the image embeddings $X_{img_{t1}}^{\prime}$ and $X_{img_{t2}}^{\prime}$ for the bitemporal images, ReasonCD uses the embedding features of all patches from the penultimate layer of the CLIP image encoder, expressed as:
% eq5
\begin{equation}
\label{eq5}
I_{img_{t1}}^{\prime}=CLIP\_Img_{Enc}(CLIP_{Process}(X_{img_{t1}}^{\prime}))[-2][1:]
\end{equation}
% eq6
\begin{equation}
\label{eq6}
I_{img_{t2}}^{\prime}=CLIP\_Img_{Enc}(CLIP_{Process}(X_{img_{t2}}^{\prime}))[-2][1:]
\end{equation}
where, $X_{img_{t1}}^{\prime},X_{img_{t2}}^{\prime}\in\mathbb{R}^{3\times H\times W}$ are the bitemporal images, $CLIP_{Process}(\bullet)$ is CLIP's image preprocessing operation, which first adjusts the mean and standard deviation of the RGB channels of the original image to [0.48145466, 0.4578275, 0.40821073] and [0.26862954, 0.26130258, 0.27577711] respectively, and then uniformly resizes the image to 224×224. $CLIP\_Img_{Enc}(\bullet)$ is the CLIP image encoder, encoding the processed image to obtain features from each layer. [-2] indicates selecting the penultimate layer feature, which has a higher dimension (1024) and stronger representational capacity. [1:] indicates taking all token embedding features except the first [CLS] token. Since CLIP's patch size is 14, the number of tokens divided is 16×16. Therefore, after processing by the above formulas, the obtained image embeddings $I_{img_{t1}}^{\prime}$ and $I_{img_{t2}}^{\prime}$ have a dimension of $\mathbb{R}^{256\times1024}$.

Since the text embedding obtained in the previous section is  $text\_token_{embedding}\in\mathbb{R}^{num\_tokens\times4096}$, $I_{img_{t1}}^{\prime}$ and $I_{img_{t2}}^{\prime}$ cannot be directly implanted. Therefore, a linear projection layer is added to map the dimension 1024 of $I_{img_{t1}}^{\prime}$ and $I_{img_{t2}}^{\prime}$ to 4096, obtaining $I_{img_{t1}}$ and $I_{img_{t2}}\in\mathbb{R}^{256\times4096}$. The tokens corresponding to $<ImgT1>$ and $<ImgT2>$ in $text\_token_{embedding}$ are replaced with $I_{img_{t1}}$ and $I_{img_{t2}}$ to obtain the final text embedding implanted with image information: $text\_img\_token_{embedding}\in\mathbb{R}^{(num\_tokens-2+256\times2)\times4096}$. Subsequently, $text\_img\_token_{embedding}$ is input into the large language model to complete the subsequent reasoning task.

\subsection{Reasoning Core: Large Language Model LLaMA2\cite{llama2}}
To reduce GPU memory usage while ensuring model reasoning capability, this paper uses LLaMA2-7b-Chat\cite{llama2} as the large language model in ReasonCD. LLaMA\cite{llama2} is a series of large language models developed by Meta AI, including four versions: 7B, 13B, 33B, and 65B. These models are trained using publicly available datasets, ensuring open-source compatibility and reproducibility. LLaMA\cite{llama} models perform excellently under various inference budgets, especially their ability to run on a single GPU, making them an efficient language model.

Regarding model training, compared to LLaMA1, LLaMA2's data is expanded by 40\%, with its training dataset reaching 2 trillion tokens, significantly expanding its vocabulary range. Additionally, LLaMA2 extends the context understanding length from the original 2048 tokens in LLaMA models to 4096 tokens. LLaMA2-7b-Chat is specifically optimized for dialogue scenarios based on LLaMA2-7b. After supervised fine-tuning (SFT) and reinforcement learning from human feedback (RLHF), it better meets human demands for usefulness and safety. In the SFT phase, the model is trained on human-annotated instruction data to improve its performance on specific tasks. In the RLHF phase, the model is further optimized for dialogue generation tasks by combining reinforcement learning with human feedback. In human evaluations, its dialogue generation capability is comparable to some popular large closed-source models (e.g., ChatGPT\cite{chatgpt} and PaLM\cite{palm}).

Regarding model architecture, although LLaMA2 still scales the model by stacking Transformers, compared to the standard Transformer, as shown in Figure \ref{fig4}, improvements in LLaMA2 include:
\begin{figure}%[h]  Figure 4
    \centering
    \includegraphics[width=0.5\textwidth]{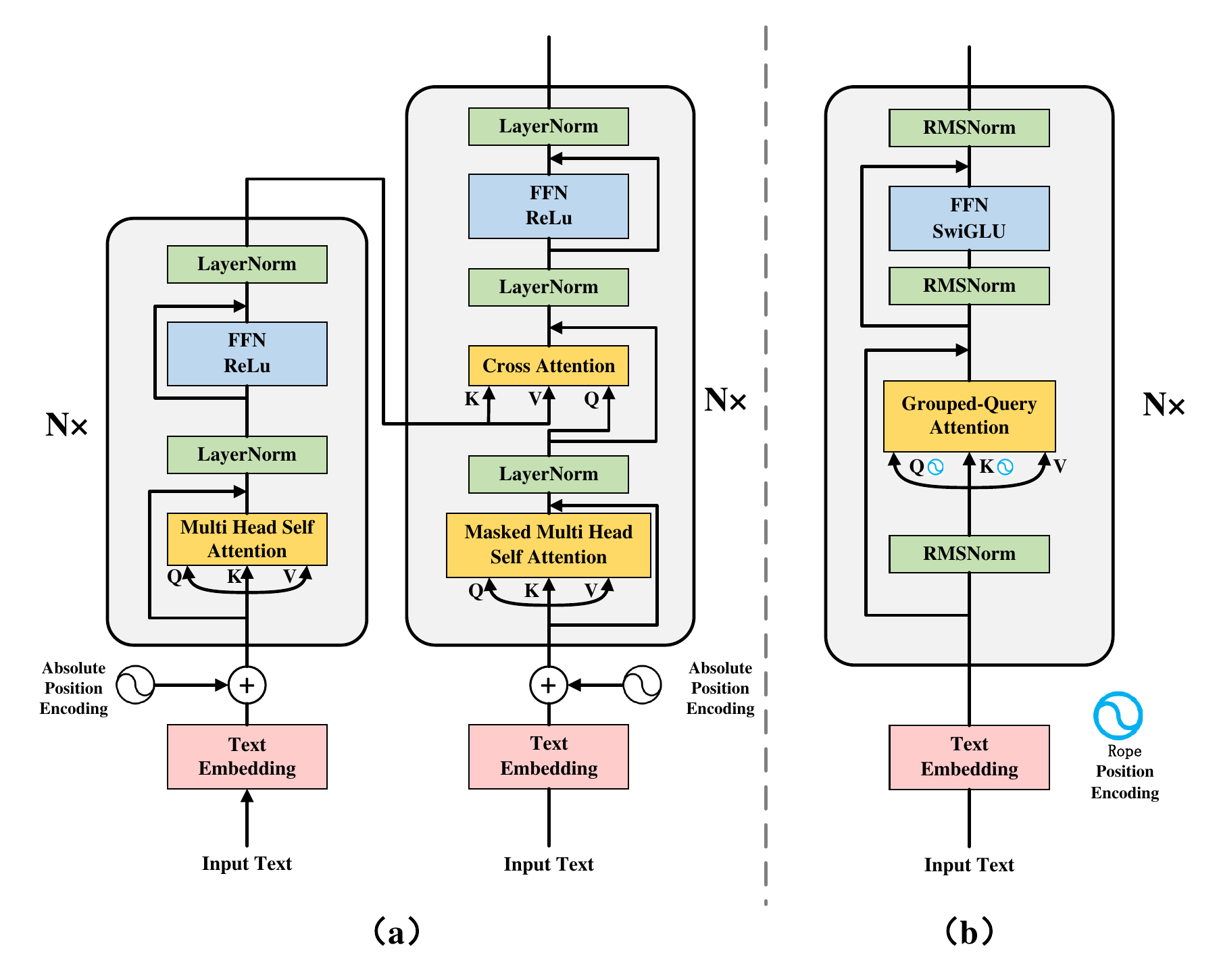}
    \caption{Comparison of Transformer architectures between (a) the standard Transformer and (b) LLaMA2.}
    \label{fig4}
\end{figure}
\subsubsection{\textbf{Adoption of Decoder-Only Architecture}}
LLaMA2 removes the encoder of the standard Transformer, expanding model parameters solely by stacking Transformer decoders. The advantages are: Theoretically, the Decoder-Only architecture combined with the next-token prediction pretraining task exposes each position to less information, making predicting the next token more difficult. This implies that when the model is sufficiently large or data is abundant, this architecture can learn a higher upper bound of general representation. Simultaneously, this architecture aligns well with the characteristics of generative tasks, where the model needs to construct the output sequence step-by-step based on previously generated content. This requires the model to effectively utilize prior information to maintain output consistency and coherence. The unidirectional causal attention, focusing only on the already generated parts, ensures the attention matrix is a lower triangular matrix, guaranteeing full rank and enhancing modeling capability.

In engineering practice, the Decoder-Only architecture omits the encoder part, improving model training and inference speed, especially prominent in autoregressive tasks. Moreover, the Decoder-Only architecture can build deeper networks with the same parameter count, thereby increasing model capacity and performance.
\subsubsection{\textbf{Replacement of LayerNorm\cite{layernorm} with RMSNorm\cite{rmsnorm}}}
The normalization method in the standard Transformer is LayerNorm, which normalizes across the feature dimension. Its calculation formula is:
% eq7
\begin{equation}
\label{eq7}
LayerNorm(x)=\frac{x-E(x)}{\sqrt{Var(x)+\varepsilon}}*\gamma+\beta
\end{equation}
where $E(x)$ and $Var(x)$ represent the mean and variance of input x across the feature dimension, respectively. $\varepsilon$ is a very small number to ensure the denominator is not zero. $\gamma$ and $\beta$ are two learnable parameters for scaling and shifting, respectively.

RMSNorm improves upon LayerNorm by removing the centering operation, which can be seen as a special case of LayerNorm with a mean of zero, avoiding mean calculation and improving computational efficiency. Its calculation formula is:
% eq8
\begin{equation}
\label{eq8}
RmsNorm(x)=\frac{x}{Rms(x)+\varepsilon}*\gamma
\end{equation}
In some applications, LayerNorm may cause training instability due to unstable mean calculation (e.g., large bias or noise). RMSNorm reduces this instability by removing the mean calculation.
\subsubsection{\textbf{Replacement of Multi-Head Attention (MHA)\cite{transformer} with Grouped-Query Attention (GQA)\cite{gqa}}}
\begin{figure}%[h]  Figure 5
    \centering
    \includegraphics[width=0.5\textwidth]{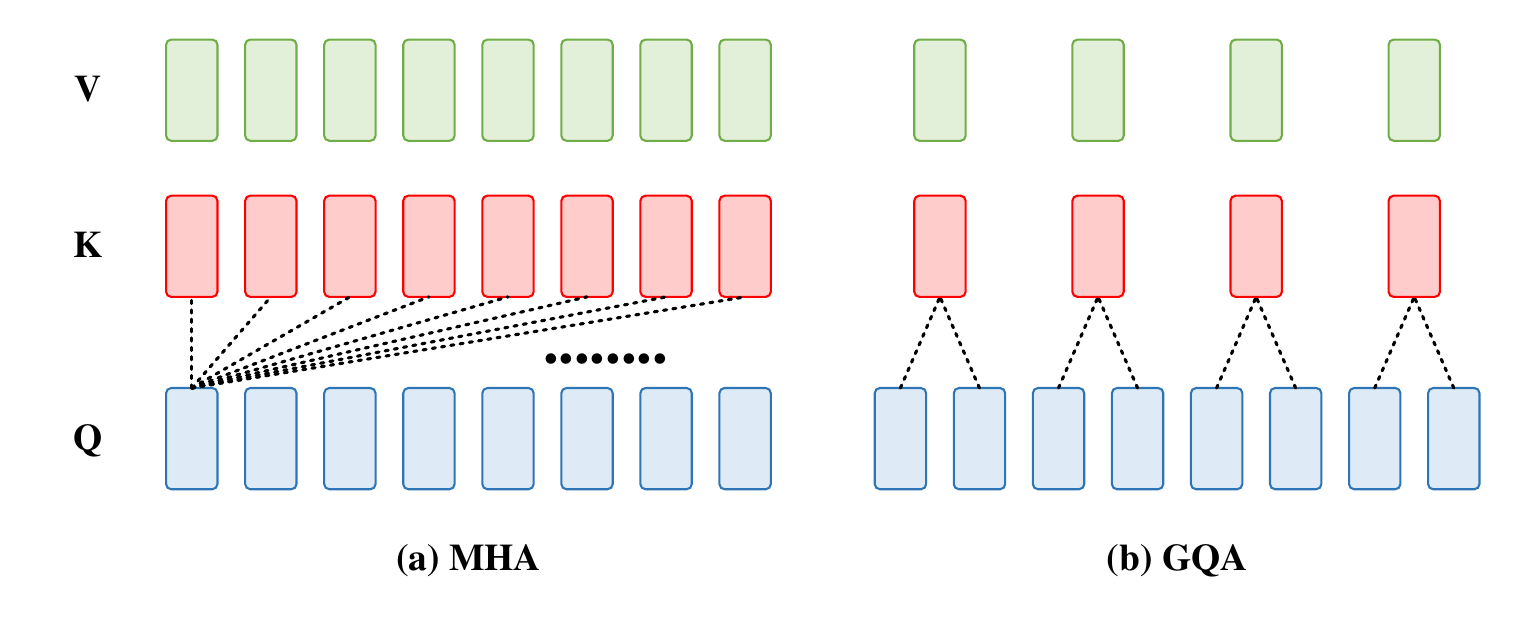}
    \caption{Comparison diagram of (a) MHA and (b) GQA structures.}
    \label{fig5}
\end{figure}
As shown in Figure \ref{fig5}(a), in MHA, the number of K and V matrices is usually equal to Q, and they correspond one-to-one. When computing attention, each Q is dot-multiplied with all Ks, and finally, the results of all Qs are aggregated for output. As shown in Figure 5(b), GQA first groups Q, with each group sharing a K and V. When computing attention, GQA only performs dot multiplication between Q and K within the same group.

Therefore, compared to MHA, GQA reduces the number of K and V matrices, lowering memory usage, allowing processing of longer sequences or larger batches under the same computational resources. Simultaneously, GQA allows multiple queries to share the same set of keys and values, reducing computational redundancy among query heads, thereby lowering computational complexity and improving inference speed.
\subsubsection{\textbf{Replacement of Absolute Positional Encoding\cite{transformer} with Rotary Positional Encoding (RoPE)\cite{rope}}}
LLaMA chooses Rotary Positional Encoding (RoPE) primarily because RoPE can more naturally capture relative positional information between tokens while having better long-sequence extrapolation capability, whereas absolute positional encoding is often limited by the training sequence length.

Also, as shown in Figure \ref{fig4}, in LLaMA, RoPE is applied only to Q and K, not directly to the original token embeddings. The purpose is to allow the dot product operation in self-attention to naturally capture relative positional information between tokens while decoupling semantic information and positional encoding in Q and K through separate application, maintaining the stability of token semantics.
\subsubsection{\textbf{Replacement of ReLU with SwiGLU\cite{swiglu} as Activation Function in Feed-Forward Network}}
SwiGLU combines the smooth characteristics of the Swish function with a gating mechanism. Compared to ReLU, which merely truncates negative values, SwiGLU can more flexibly capture complex nonlinear relationships, thereby enhancing network expressiveness. Secondly, ReLU has zero gradient when input is negative, potentially causing the "dying neuron" problem. SwiGLU, due to its smooth nature and gating mechanism, can maintain more continuous gradients, helping alleviate the vanishing gradient problem and improving training stability and convergence speed.
\subsubsection{\textbf{Replacement of Post-Layer Normalization with Pre-Layer Normalization}}
As also seen in Figure \ref{fig4} compared to the traditional Transformer's post-layer normalization (i.e., normalization after residual connection), LLaMA adopts a pre-normalization strategy (normalization before residual connection) to enhance training stability and optimize gradient flow.

Traditional post-normalization can easily cause vanishing or exploding gradient problems in deep Transformer models because the normalization operation is performed after the residual connection, and the residual signal may be weakened. Pre-normalization normalizes the input before each sub-layer, ensuring stable input distribution for each layer, thereby making gradient propagation smoother.

Finally, this paper uses a Greedy Search\cite{greedy_search} strategy to sample the output of LLaMA, i.e., taking the token with the highest probability as the result for each generation step.
\subsection{LoRA Efficient Fine-Tuning}
LoRA is a commonly used efficient fine-tuning technique for large models. Its basic idea is to constrain the weight matrix update as the product of two low-rank matrices, thereby reducing the number of parameters that need to be learned.
\begin{figure}%[h]  Figure 6
    \centering
    \includegraphics[width=0.3\textwidth]{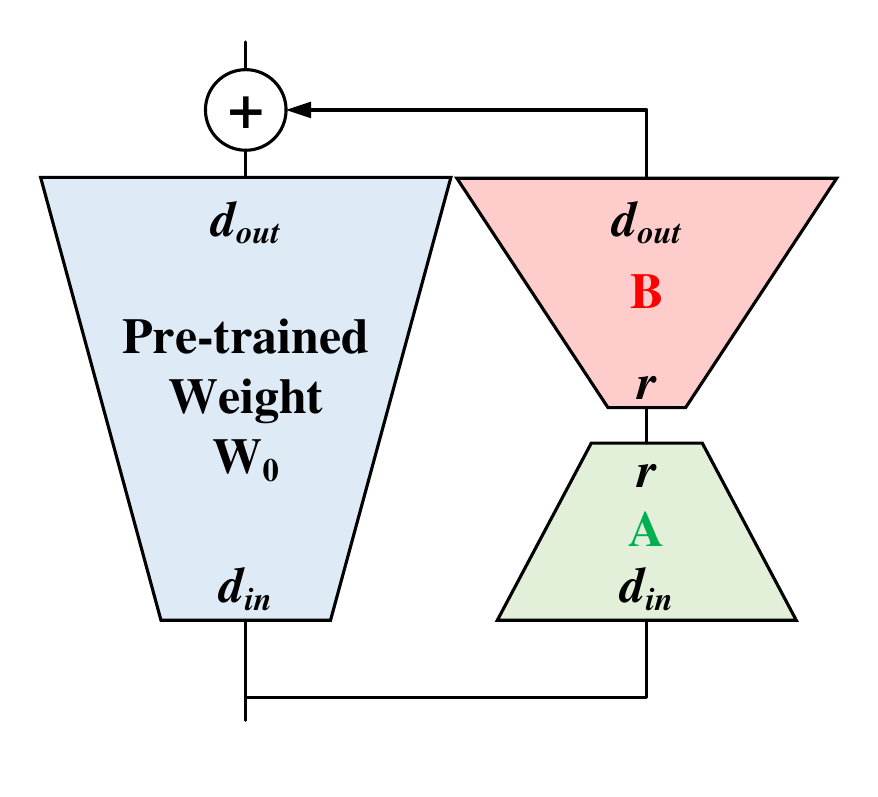}
    \caption{LoRA Structure Diagram.}
    \label{fig6}
\end{figure}
Figure \ref{fig6} illustrates the working principle of LoRA. Assuming the pre-trained original model weight matrix is $W_0$, LoRA adds a low-rank update matrix $\Delta W$ to it, forming a new weight matrix $W$:
% eq9
\begin{equation}
\label{eq9}
W=W_0+\Delta W
\end{equation}
where $W_0$ is obtained by multiplying the two low-rank matrices $A$ and $B$ in Figure \ref{fig6}:
% eq10
\begin{equation}
\label{eq10}
\Delta W = A \times B
\end{equation}
Assuming the shape of $W_0$ is $(d_{in}\times d_{out})$ , then matrix $A$ has shape $(d_{in}\times r)$, and matrix $B$ has shape $(r\times d_{out})$, where $(d_{in}$ and $(d_{out}$ epresent the input and output dimensions of this matrix, respectively, and $r$ is the set low rank. During training, only $A$ and $B$ are updated, while $W_0$ remains frozen. Therefore, compared to full-parameter fine-tuning, the number of learnable parameters reduced using LoRA for fine-tuning is $d_{in}\times d_{out} - d_{in}\times r - r\times d_{out}$. Thus, when $r$ is much smaller than $(d_{in}$ and $(d_{out}$, the number of learnable parameters is significantly reduced, improving computational efficiency.

Parameters of matrix $A$ are initialized using Kaiming initialization\cite{kaiming_init}, while parameters of matrix $B$ are initialized to 0.  The reasons for initializing matrix $B$ to 0 can be summarized in three points:
\subsubsection{\textbf{Gradual Introduction of LoRA Weight Adjustments}}
With $B$ initialized to 0, $A\times B=0$, meaning LoRA adjustments do not take effect immediately, and the model relies entirely on the pre-trained weights $W_0$. This helps the model gradually adapt to LoRA adjustments, preventing the model from overly depending on LoRA-modified weights from the start.
\subsubsection{\textbf{Stable Training Dynamics}}
In the initial stage, zero initialization of matrix $B$ reduces disturbance from LoRA, making the model rely on the original pre-trained weights. During training, as matrix $B$ is updated, the effects of LoRA adjustments accumulate gradually. This prevents excessive weight changes early in training, avoiding unnecessary gradient explosion or instability.
\subsubsection{\textbf{Control of Weight Update Magnitude}}
Initializing matrix $B$ to zero effectively reduces the magnitude of weight updates in the early training stage, making the model's learning process more stable and gentle. This strategy is particularly effective in scenarios with small batch sizes or where the model might overfit initially.

Matrix $A$ cannot be initialized to 0 because it would prevent effective gradient updates, and the model could not learn effective low-rank adjustments. Matrix $A$ needs to participate in learning from the beginning to ensure effective weight adjustment by LoRA.

\subsection{Multi-scale Feature Fusion Module $fu(\bullet)$}
ReasonCD's ChangeMap Decoder receives a high-dimensional semantic feature prompt from the $<CHG>$ embedding and also receives a feature $fu({X_{img_{t1}},X_{img_{t2}})}\in\mathbb{R}^{256\times64\times64}$ that fuses the bitemporal images. This section focuses on the process of obtaining this feature.

Since the ChangeMap Decoder is the mask decoder of SAM\cite{sam}, using SAM's image encoder can obtain features more compatible with it. However, SAM's image encoder stacks a large number of Transformer blocks, has deep network layers, and tends to extract high-level semantic features, lacking in capturing low-level visual features.

\begin{figure*}[ht]%[h]  Figure 7
\centering
\includegraphics[width=7in]{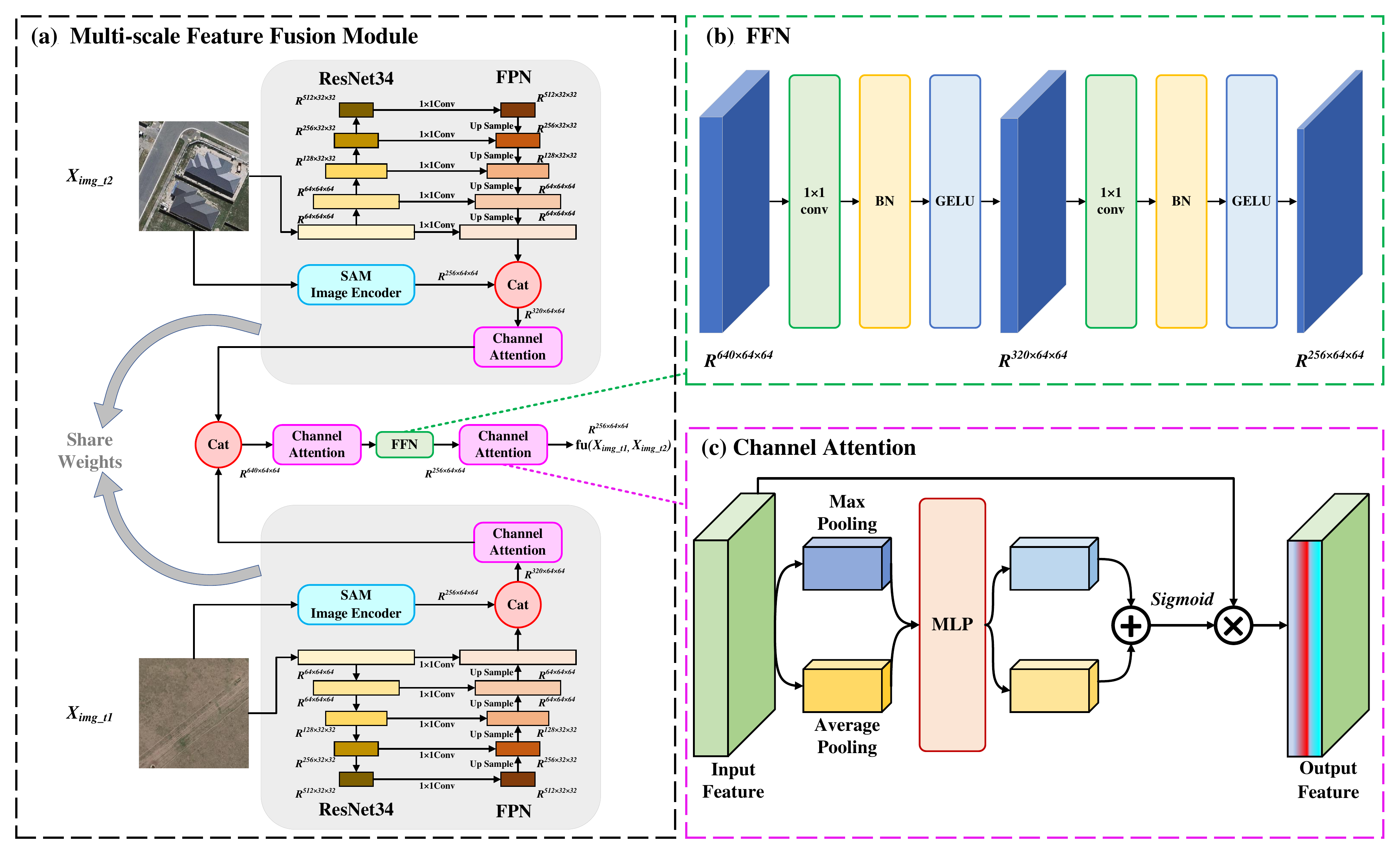}
\caption{Schematic Diagram of the Multi-scale Feature Fusion Module $fu(\bullet)$.}
\label{fig7}
\end{figure*}

Therefore, in the Multi-scale Feature Fusion Module (as shown in Figure \ref{fig7}(a)), we use ResNet34\cite{resnet} and FPN\cite{fpn} to extract low-level visual features from the images. These are concatenated with features extracted by the SAM image encoder. A channel attention mechanism\cite{channel_attention} is used to weight the high-level and low-level features, automatically capturing useful information fused from both. The bitemporal images undergo feature extraction in the same manner. After concatenating the feature maps, channel attention and a Feed-Forward Network (FFN) are used to map and transform the features, ultimately producing the output $fu({X_{img_{t1}},X_{img_{t2}})}$.

% \begin{figure}%[h]  Figure 7
%     \centering
%     \includegraphics[width=0.5\textwidth]{fig/fig7.pdf}
%     \caption{Schematic Diagram of the Multi-scale Feature Fusion Module $fu(\bullet)$}
%     \label{fig7}
% \end{figure}

The structure of the Feed-Forward Network FFN is shown in Figure \ref{fig7}(b), mainly consisting of two layers of 1×1 convolutional networks used primarily for channel transformation. Each convolution operation is followed by a Batch Normalization (BN) layer and a GELU activation function layer\cite{gelu}. The structure of the channel attention is shown in Figure \ref{fig7}(c). It splits the input features into two branches: an average pooling path and a max pooling path. Then, a shared-weight Multi-Layer Perceptron (MLP) performs nonlinear transformation on the pooled features. The nonlinearly transformed features from the two branches are added, and a Sigmoid function is used to calculate the channel attention scores. Finally, the original input features are multiplied by the channel attention scores to obtain the final output features. 

\subsection{Loss Functions Used in Training}
ReasonCD uses an end-to-end training approach. The training objective primarily optimizes two types of losses: text generation loss $L_{text}$ and change mask loss $L_{cd}$.

The text generation loss is mainly the loss between the text generated by the large language model LLaMA2 and the annotated text. Since ReasonCD uses a greedy search strategy, this paper uses weighted cross-entropy loss as the text generation loss:
% eq11
\begin{equation}
\label{eq11}
L_{text}=\lambda_{ce}L_{ce}
\end{equation}
where $L_{ce}$ is the cross-entropy loss function, calculated as:
% eq12
\begin{equation}
\label{eq12}
L_{ce}(y_i,y_i^{\prime})=\sum_{i=1}^ny_i\log y_i^{\prime}
\end{equation}
Here, $y_i$ is the one-hot encoding of the ground truth label, and $y_i^{\prime}$ is the probability distribution predicted by the model. Additionally, considering that the corresponding features for subsequent change map generation can only be extracted when the $<CHG>$ special token appears in the output text, the loss penalty is increased when the $<CHG>$ token does not appear in the generated text. This paper implements this process through the weight $\lambda_{ce}$ of $L_{ce}$.Specifically, when the $<CHG>$ token does not appear in the generated text, the loss weight corresponding to the $<CHG>$ token category becomes 10 times the original.

The change mask loss is the pixel-level loss for the final generated change map. During training, this paper selects weighted binary cross-entropy loss $L_{bce}$ and  DICE loss\cite{dice} $L_{dice}$:
% eq13
\begin{equation}
\label{eq13}
L_{cd}=\lambda_{bce}L_{bce} + \lambda_{dice}L_{dice}
\end{equation}
Assuming the final predicted change probability map by the ChangeMap Decoder is $M^{\prime}$, , and the annotated corresponding change map is $M$, the formulas for $L_{bce}$ and  $L_{dice}$ are respectively:
% eq14
\begin{equation}
\label{eq14}
L_{_{bce}}(M,M^{\prime})=M\log M^{\prime}+(1-M)\log(1-M^{\prime})
\end{equation}
% eq15
\begin{equation}
\label{eq15}
L_{dice}\left(M,M^{\prime}\right)=1-\frac{2\sum\left(M^{\prime}\cdot M\right)+\varepsilon}{\sum M^{\prime}+\sum M+\varepsilon}
\end{equation}
Therefore, the total loss $L$ for optimizing ReasonCD is:
% eq16
\begin{equation}
\begin{aligned}
\label{eq16}
L&=L_{text}+L_{cd}\\&=\lambda_{ce}L_{ce}\left(y_{text},y_{text}^{\prime}\right)+\lambda_{bce}L_{bce}\left(M,M^{\prime}\right)\\&+\lambda_{dice}L_{dice}\left(M,M^{\prime}\right)
\end{aligned}
\end{equation}
where $y_{text}$, $y_{text}^{\prime}$ represent the probability of the generated text and the label for the generated text, respectively.

\section{Experiments and discussion}
\subsection{Datasets}
To verify the feasibility of ReasonCD and its performance in implicit semantic reasoning, this paper uses the BCDD dataset\cite{bcdd} and the SECOND dataset\cite{second}. The BCDD dataset is primarily used for subsequent model accuracy comparison. Additionally, since the SECOND dataset is a multi-category change detection dataset, this paper annotates some validation data for implicit semantic reasoning based on its 6 categories.

\begin{figure}%[h]  Figure 8
    \centering
    \includegraphics[width=0.5\textwidth]{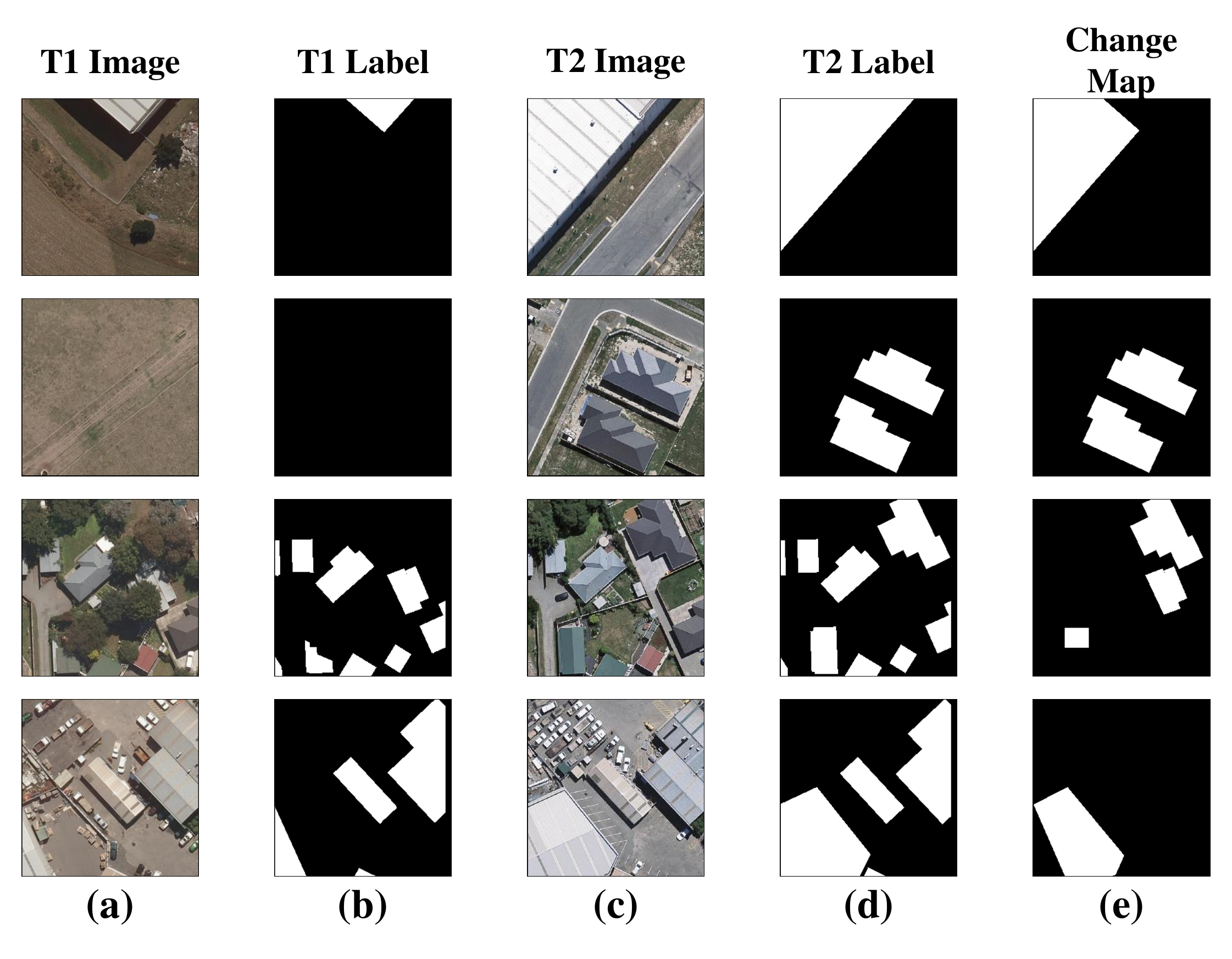}
    \caption{Multi temporal image samples selected from the BCDD dataset.}
    \label{fig8}
\end{figure}

As shown in Figure \ref{fig8}, the BCDD dataset covers areas affected by the 6.3 magnitude earthquake that struck Christchurch, New Zealand, in February 2011. The dataset contains two image scenes acquired at the same location in 2012 and 2016, along with building semantic labels and change labels, aiming to study the detection and analysis of building changes post-disaster. Considering the original image size is 32,507 × 15,354 pixels, this paper employs a tiling method for the large images. Specifically, each large image is split into non-overlapping 256 × 256 pixel small image pairs, effectively reducing the computational burden of data processing. After tiling, a total of 7,434 bitemporal image pairs are obtained for subsequent change detection tasks.

\begin{figure}%[h]  Figure 9
    \centering
    \includegraphics[width=0.5\textwidth]{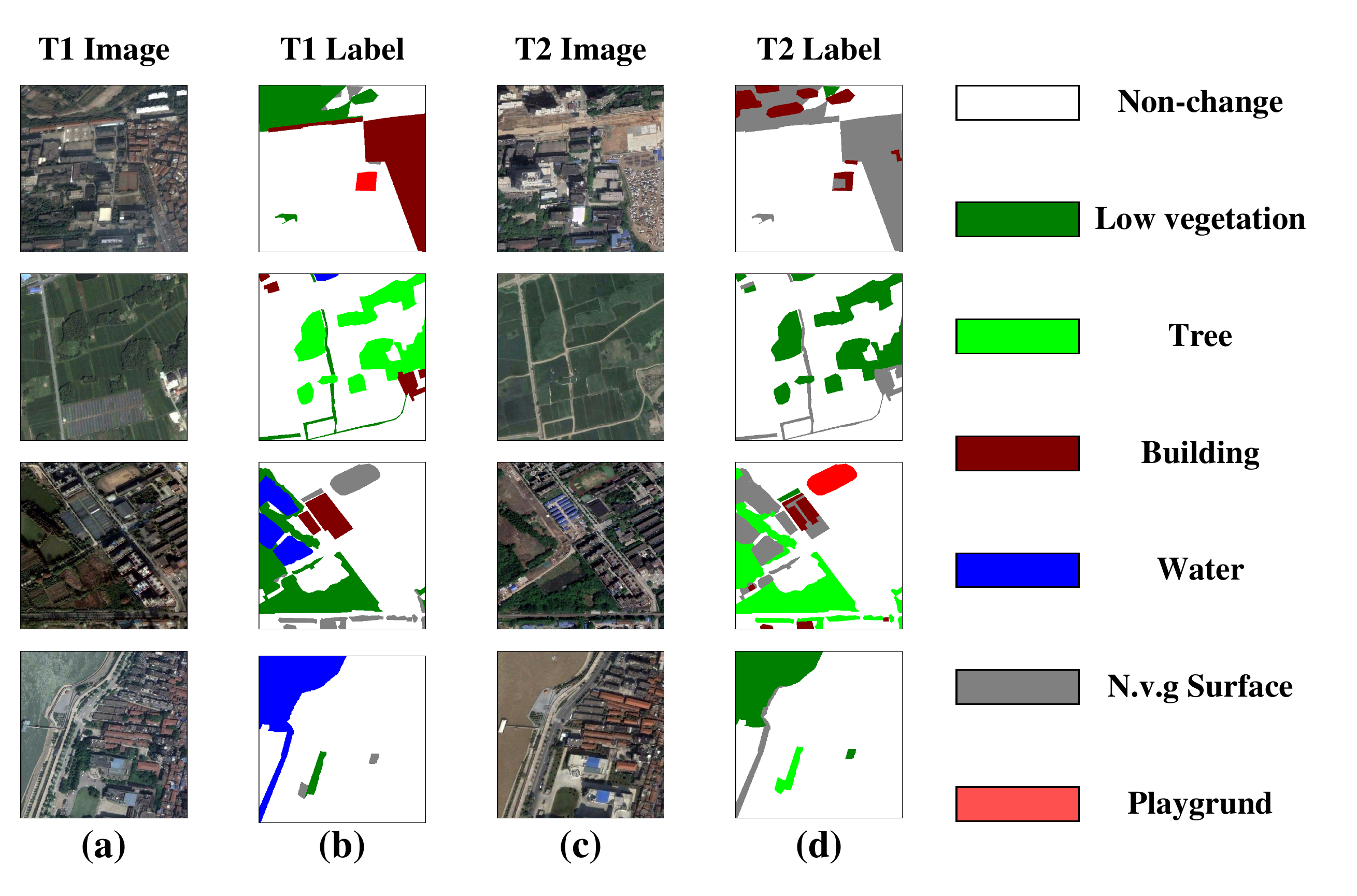}
    \caption{Multi temporal image samples selected from the SECOND dataset.}
    \label{fig9}
\end{figure}

As shown in Figure \ref{fig9}, the SECOND dataset is a public dataset for semantic change detection tasks. It contains 4,662 pairs of aerial images from various platforms and sensors covering multiple cities, including Hangzhou, Chengdu, and Shanghai. Each image is 512 × 512 pixels and is annotated in detail at the pixel level. The SECOND dataset focuses on six land cover types: non-vegetated surface, trees, low vegetation, water, buildings, and playground. For the change detection task, this paper establishes six CRoI baselines based on the land cover types at time T1. The dataset is thus divided into six CRoI categories for more fine-grained change detection analysis.

Furthermore, to more fully utilize the datasets and adapt to ReasonCD's input, this paper reconstructs the dataset storage format, storing each training sample using a JSON file. The JSON file structure is shown in Table \ref{tab2}.
\begin{table}[]
\centering
\caption{JSON File Structure for ReasonCD Training}
\label{tab2}
\resizebox{\columnwidth}{!}{%
\begin{tabular}{@{}cc@{}}
\toprule
JSON Field     & Meaning 
\\ \midrule
"img1\_path" & Path to T1 image \\
"img2\_path" & Path to T2 image \\
"mask\_path" & Path to the corresponding CRoI change map mask \\
"question" & Implicit CRoI description requiring reasoning \\
"answer" & Annotated text generated by the large language model \\
\bottomrule
\end{tabular}%
}
\end{table}
For a training sample's JSON file, it contains 5 fields: "img1\_path", "img2\_path", "mask\_path", "question", and "answer", representing the path to the T1 image, path to the T2 image, path to the corresponding CRoI change map mask, the implicit CRoI description requiring reasoning, and the annotated text generated by the large language model, respectively. Since LLaMA2 has a larger English corpus, all textual annotations in the data use English to accelerate model convergence.

Regarding dataset usage, this paper further divides it into training, validation, and test sets with a ratio of 8:1:1, ensuring the scientificity and effectiveness of model training and evaluation.

\subsection{Metrics and Implementation Details}
To quantitatively study the performance of various change detection methods, this paper uses commonly used change detection metrics: F1 Score (F1), Recall, Precision, Overall Accuracy (OA), Kappa Coefficient (Kappa), and Intersection over Union (IoU). Their calculation formulas are as follows:
% eq17
\begin{equation}
\label{eq17}
Precision=\frac{TP}{TP+FP}
\end{equation}
% eq18
\begin{equation}
\label{eq18}
Recall=\frac{TP}{TP+FN}
\end{equation}
% eq19
\begin{equation}
\label{eq19}
F1=\frac{2\times Precsion\times Recall}{Precsion + Recall}
\end{equation}
% eq20
\begin{equation}
\label{eq20}
OA=\frac{TP+TN}{TP+TN+FP+FN}
\end{equation}
% eq21
\begin{equation}
\label{eq21}
IoU=\frac{TP}{TP+TN+FP}
\end{equation}
% eq22
\begin{equation}
\label{eq22}
Kappa=\frac{p_0-p_e}{1-p_e}
\end{equation}
% eq23
\begin{equation}
\label{eq23}
p_0=\frac{TP+TN}{TP+TN+FP+FN}
\end{equation}
% eq24
\begin{equation}
\begin{aligned}
\label{eq24}
p_e&=\\&\frac{(TP+FP)\times(TP+FN)+(TN+FN)+(TN+FP)}{\left(TP+TN+FP+FN\right)^2}
\end{aligned}
\end{equation}
where TP is the number of true positives, FP is the number of false positives, TN is the number of true negatives, and FN is the number of false negatives.

All algorithms and models in the experiments are implemented using the PyTorch backend, with computational support from an NVIDIA RTX A6000 GPU (48GB). The CPU model is Intel(R) Xeon(R) Gold 6248R CPU @ 3.00GHz.

\subsection{Model Performance Comparison Experiment}
Since no current work focuses on implicit CRoI reasoning change detection methods, to validate ReasonCD's change detection effectiveness, this experiment compares ReasonCD with current advanced supervised change detection baselines on the BCDD test set for building change detection metrics. Experimental details and results are as follows:

For supervised baselines, this experiment uses 8 advanced methods for comparison: CDNet \cite{cdnet}, FC-EF\cite{fcef}, FC-Siam-conc\cite{fcef}, FC-Siam-diff\cite{fcef}, BiDateNet\cite{bidatenet}, DASNet\cite{dasnet}, EGCTNet\cite{egctnet}, HFA-Net\cite{hfanet}. All methods use their optimal parameters as set in their respective papers to ensure maximum performance.

For ReasonCD, the frozen parameter parts load corresponding official pre-trained weights. The initial learning rate is set to 0.00005, the optimizer is AdamW\cite{adamw}, and the learning rate policy uses a warm-up followed by decay strategy, with a warm-up period of 100 steps. The batch size is set to 16, and the total training epochs are set to 100, but early stopping is applied based on overfitting conditions to prevent overfitting and save resources. The three loss weights for training optimization, $\lambda_{ce}$,$\lambda_{bce}$ and $\lambda_{dice}$ are set to 1.0, 2.0 and 0.5. Training precision is float32, and the DeepSpeed library\cite{deepspeed} is used for acceleration. To reduce training GPU memory usage, this experiment uses LoRA to fine-tune the large language model LLaMA2 in ReasonCD. The LoRA rank is set to 8, primarily applying low-rank adaptation to the q and v matrices of each self-attention layer in LLaMA2. Furthermore, since the task is building change detection, to disregard performance drops caused by implicit CRoI reasoning, when preparing JSON files, the "question" field's implicit CRoI text is uniformly set to "Building", and the "answer" field is uniformly set to "It's $<CHG>$".

The performance metrics of ReasonCD and the aforementioned 8 supervised baseline methods on the BCDD test set are shown in Table \ref{tab3}. The experimental results show that except for the Precision metric, ReasonCD outperforms all 8 supervised baselines in other metrics, with an F1 score reaching 0.921. This sufficiently demonstrates that ReasonCD achieves good detection performance even in the simple single-CRoI fitting task, validating the effectiveness and feasibility of ReasonCD for change detection tasks. Notably, the ability of ReasonCD to perform implicit CRoI semantic mining is a capability these other methods lack, which will be demonstrated in subsequent experiments.
\begin{table}[]
\centering
\caption{Performance Metrics of ReasonCD and Other Supervised Baselines on the BCDD Test Set}
\label{tab3}
\resizebox{\columnwidth}{!}{%
\begin{tabular}{@{}ccccccc@{}}
\toprule
Method     & F1  & Precision & Recall & OA & IoU & Kappa
\\ \midrule
CDNet\cite{cdnet} & 0.862 & 0.908 & 0.821 & 0.989 & - & - \\
FC-EF\cite{fcef} & 0.690 & 0.690 & 0.658 & 0.944 & 0.566 & 0.624 \\
FC-Siam-Diff\cite{fcef} & 0.712 & 0.908 & 0.821 & 0.989 & - & - \\
FC-Siam-Conc\cite{fcef} & 0.712 & 0.745 & 0.739 & 0.949 & 0.601 & 0.664 \\
BiDateNet\cite{bidatenet} & 0.852 & 0.889 & 0.819 & 0.986 & - & - \\
DASNet\cite{dasnet} & 0.898 & 0.892 & 0.905 & 0.990 & - & - \\
EGCTNet\cite{egctnet} & 0.904 & \textbf{0.943} & 0.867 & 0.977 & 0.824 & 0.891 \\
HFA-Net\cite{hfanet} & 0.745 & 0.922 & 0.626 & 0.947 & 0.594 & 0.717 \\
ReasonCD(ours) & \textbf{0.921} & 0.896 & \textbf{0.947} & \textbf{0.993} & \textbf{0.853} & \textbf{0.916} \\
\bottomrule
\end{tabular}%
}
\end{table}

\subsection{Ablation Experiment for the Multi-scale Feature Fusion Module $fu(\bullet)$}
In ReasonCD, this paper designs the Multi-scale Feature Fusion Module $fu(\bullet)$ to address the issue where SAM's image encoder tends to extract higher-level semantic features while lacking low-level visual features. To validate the effectiveness of this module, the specific experimental setup is as follows. Experimental details and results are as follows:

For ReasonCD, this experiment compares the change detection metrics on the BCDD test set with and without the Multi-scale Feature Fusion Module. It also plots the training loss curves and validation metric curves for both cases. After removing the Multi-scale Feature Fusion Module, to ensure the same input feature dimensions, this experiment only concatenates the bitemporal image features extracted by SAM along the channel dimension and then uses the Feed-Forward Network in Figure \ref{fig7}(b) for dimensional mapping to be used by the subsequent ChangeMap Decoder for decoding the change map. The hyperparameters for both training configurations remain consistent with those in Subsection C.

\begin{figure}%[h]  Figure 10
    \centering
    \includegraphics[width=0.5\textwidth]{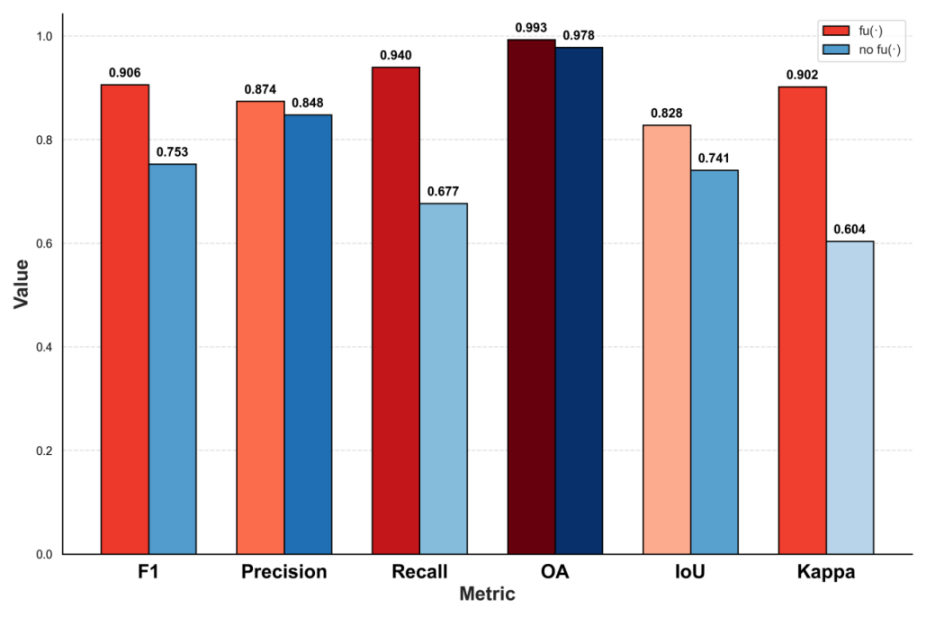}
    \caption{Bar chart of ablation experiment metrics for the Multi-scale Feature Fusion Module $fu(\bullet)$ on the BCDD test set.}
    \label{fig10}
\end{figure}

The optimal performance metrics achieved before overfitting in the ablation experiment for the Multi-scale Feature Fusion Module $fu(\bullet)$ on the BCDD dataset are shown in Figure \ref{fig10}. The red bars represent using the Multi-scale Feature Fusion Module, while the blue bars represent not using it.

The experimental results show that using the Multi-scale Feature Fusion Module leads to significant improvements across all six change detection metrics compared to not using it. The F1 score increases by 15.3 percentage points, indicating that the addition of the Multi-scale Feature Fusion Module greatly enhances ReasonCD's change detection performance.

The training loss curves on the BCDD training set for scenarios with and without the Multi-scale Feature Fusion Module $fu(\bullet)$ are shown in Figure 11. The blue curve represents the loss curve without the module, and the red curve represents the loss curve with the module.
\begin{figure}%[h]  Figure 11
    \centering
    \includegraphics[width=0.5\textwidth]{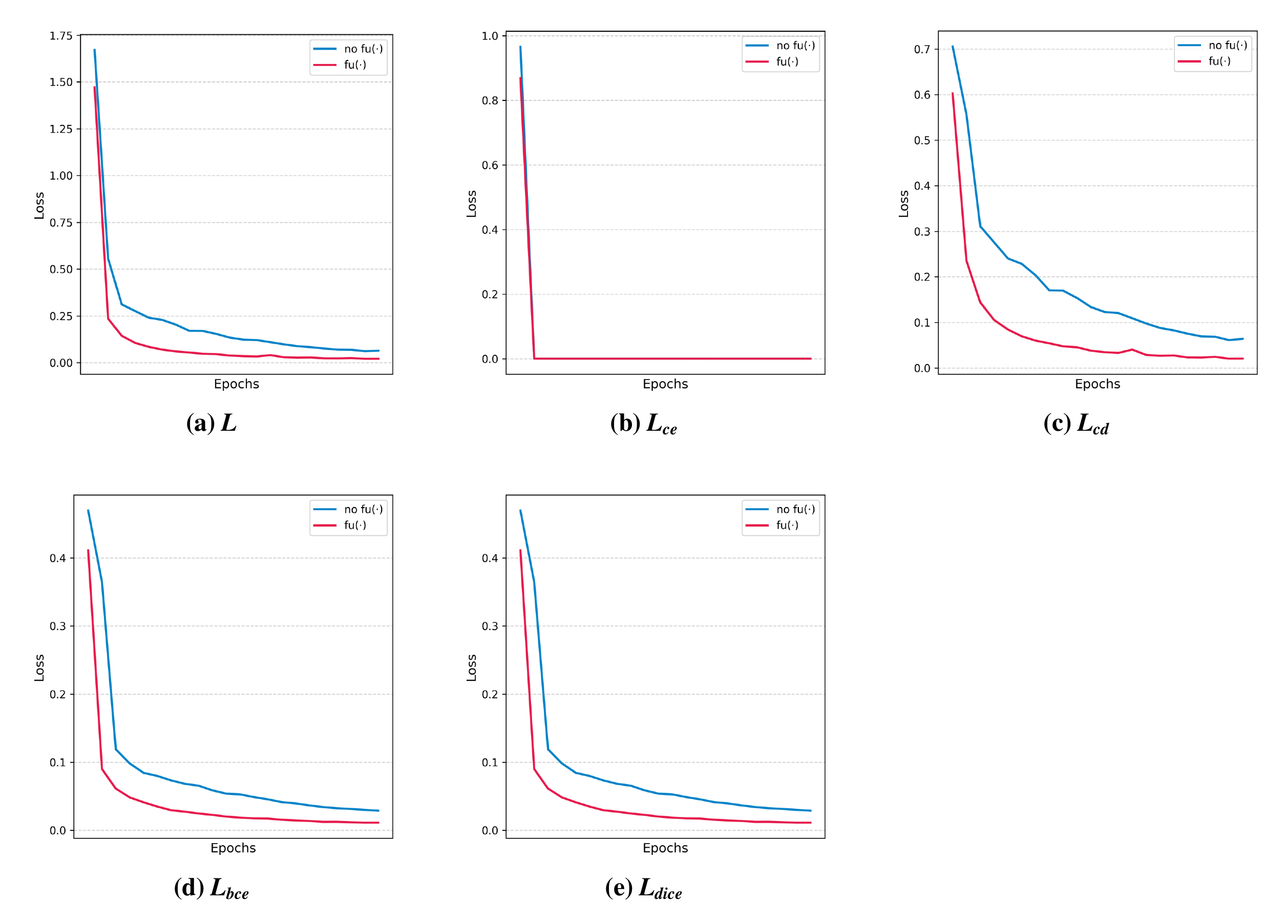}
    \caption{Training loss curves on the BCDD training set with/without the Multi-scale Feature Fusion Module $fu(\bullet)$.}
    \label{fig11}
\end{figure}

The experimental results show that in terms of the total loss $L$ (as shown in Figure \ref{fig11}(a)), using the Multi-scale Feature Fusion Module accelerates model convergence speed. Simultaneously, compared to not using the module, the Multi-scale Feature Fusion Module can optimize the loss to a lower value, enhancing the model's learning capability. Additionally, regarding the text generation loss $L_{ce}$(as shown in Figure \ref{fig11}(b)), there is almost no difference between the two, which is related to the position and function of the added module. The Multi-scale Feature Fusion Module operates after text generation and before input to the ChangeMap Decoder, primarily used for fusing high-level semantic features and low-level visual features. Therefore, it does not affect the text generation loss but acts on the change mask loss (as shown in Figures \ref{fig11}(c), (d), and (e)).

\begin{figure}%[h]  Figure 12
    \centering
    \includegraphics[width=0.5\textwidth]{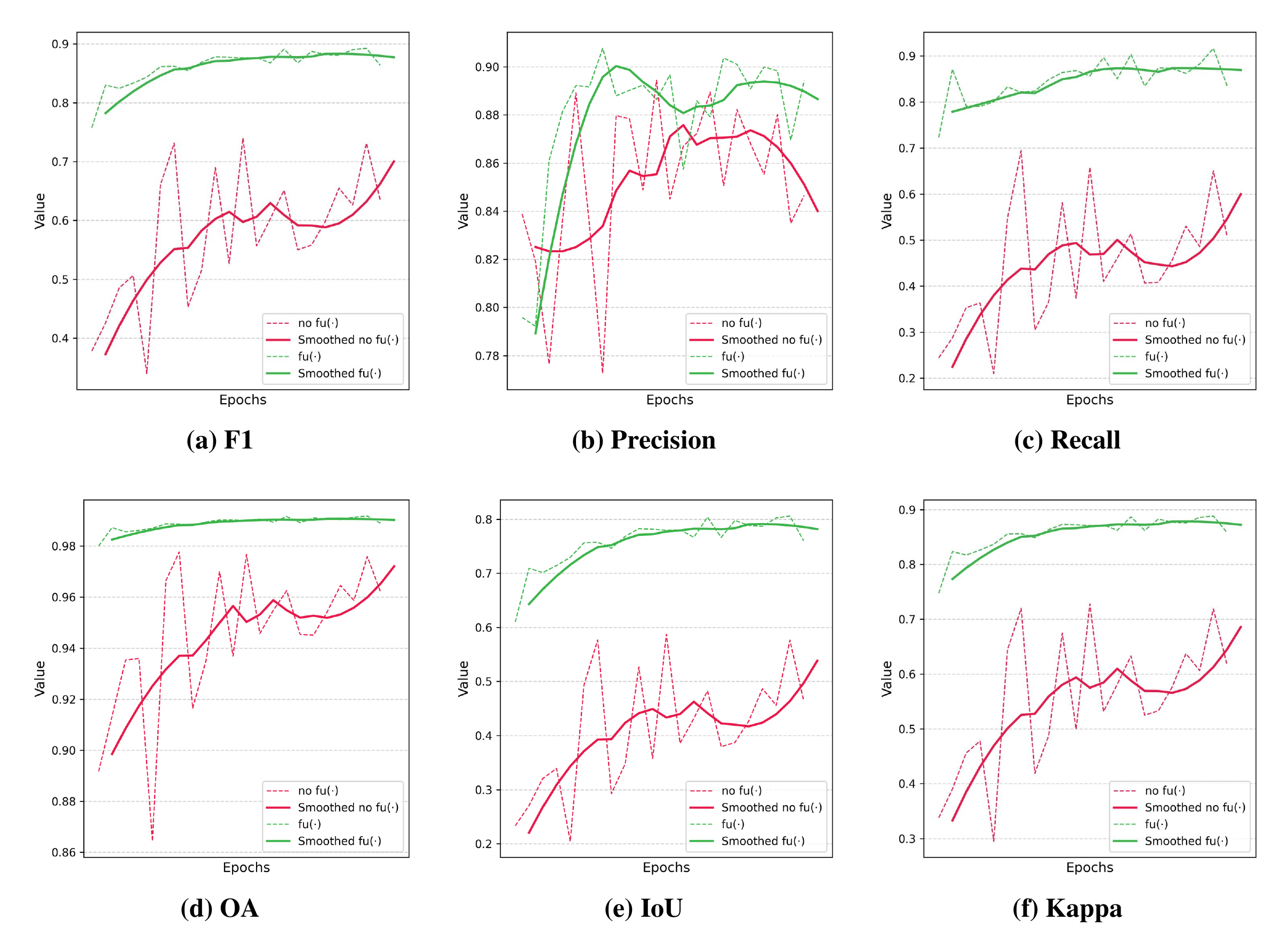}
    \caption{Training metric curves on the BCDD validation set with/without the Multi-scale Feature Fusion Module $fu(\bullet)$.}
    \label{fig12}
\end{figure}

The validation set metric curves during training on the BCDD dataset for scenarios with and without the Multi-scale Feature Fusion Module $fu(\bullet)$ are shown in Figure \ref{fig12}. The red dashed and solid lines represent the unsmoothed and smoothed metric curves, respectively, for the case without the module. The green dashed and solid lines represent the unsmoothed and smoothed metric curves, respectively, for the case with the module. 
The experimental results show that using the Multi-scale Feature Fusion Module accelerates model convergence speed, allowing the model to achieve optimal performance with fewer iterations. As can also be seen from Figure \ref{fig12}(a), after the first iteration, the F1 score with the Multi-scale Feature Fusion Module is close to 0.80, nearly 40 percentage points higher than the approximately 0.40 without the module. It is also evident that throughout the training process, the change detection metrics for ReasonCD with the Multi-scale Feature Fusion Module remain at a high level, showing a significant gap compared to the metrics without the module. Therefore, the above three experimental results validate the effectiveness of the Multi-scale Feature Fusion Module $fu(\bullet)$.

\subsection{LoRA Rank Ablation Experiment}
To reduce the GPU memory usage during ReasonCD training, this paper uses LoRA to fine-tune the large language model LLaMA2 in ReasonCD. In LoRA fine-tuning, the low rank $r$ usually needs to be set as a hyperparameter. The purpose of this experiment is to determine the low rank that yields the best change detection effect for ReasonCD and to explore its characteristics under different ranks. Experimental details and results are as follows:

For the low rank $r$ in LoRA efficient fine-tuning, this experiment sets three values: 4, 8, and 12. ReasonCD is trained under these three low ranks, and their change detection metrics on the BCDD test set are compared. Additionally, to explore the different effects brought by different ranks, this experiment also plots the training loss curves and validation metric curves under different ranks. Except for the LoRA rank, all other hyperparameters use the training parameters for ReasonCD from Subsection C.

The change detection metrics of ReasonCD under different LoRA ranks on the BCDD test set are shown in Table \ref{tab4}.
\begin{table}[]
\centering
\caption{Change Detection Metrics of ReasonCD under Different LoRA Ranks on the BCDD Test Set}
\label{tab4}
\resizebox{\columnwidth}{!}{%
\begin{tabular}{@{}ccccccc@{}}
\toprule
$r$     & F1  & Precision & Recall & OA & IoU & Kappa
\\ \midrule
$r$=4 & 0.894 & \textbf{0.897} & 0.891 & 0.992 & 0.809 & 0.890 \\
$r$=8 & \textbf{0.906} & 0.874 & \textbf{0.940} & \textbf{0.993} & \textbf{0.828} & \textbf{0.902}\\
$r$=12 & 0.888 & 0.893 & 0.884 & 0.991 & 0.79 & 0.884\\
\bottomrule
\end{tabular}%
}
\end{table}

The experimental results show that when the LoRA rank $r$ is 8, ReasonCD's metrics, except for Precision, exceed those when $r$ is 4 and 12. Therefore, it can be considered that when the rank is 8, ReasonCD's learning ability achieves a relatively optimal effect.

\begin{figure}%[h]  Figure 13
    \centering
    \includegraphics[width=0.5\textwidth]{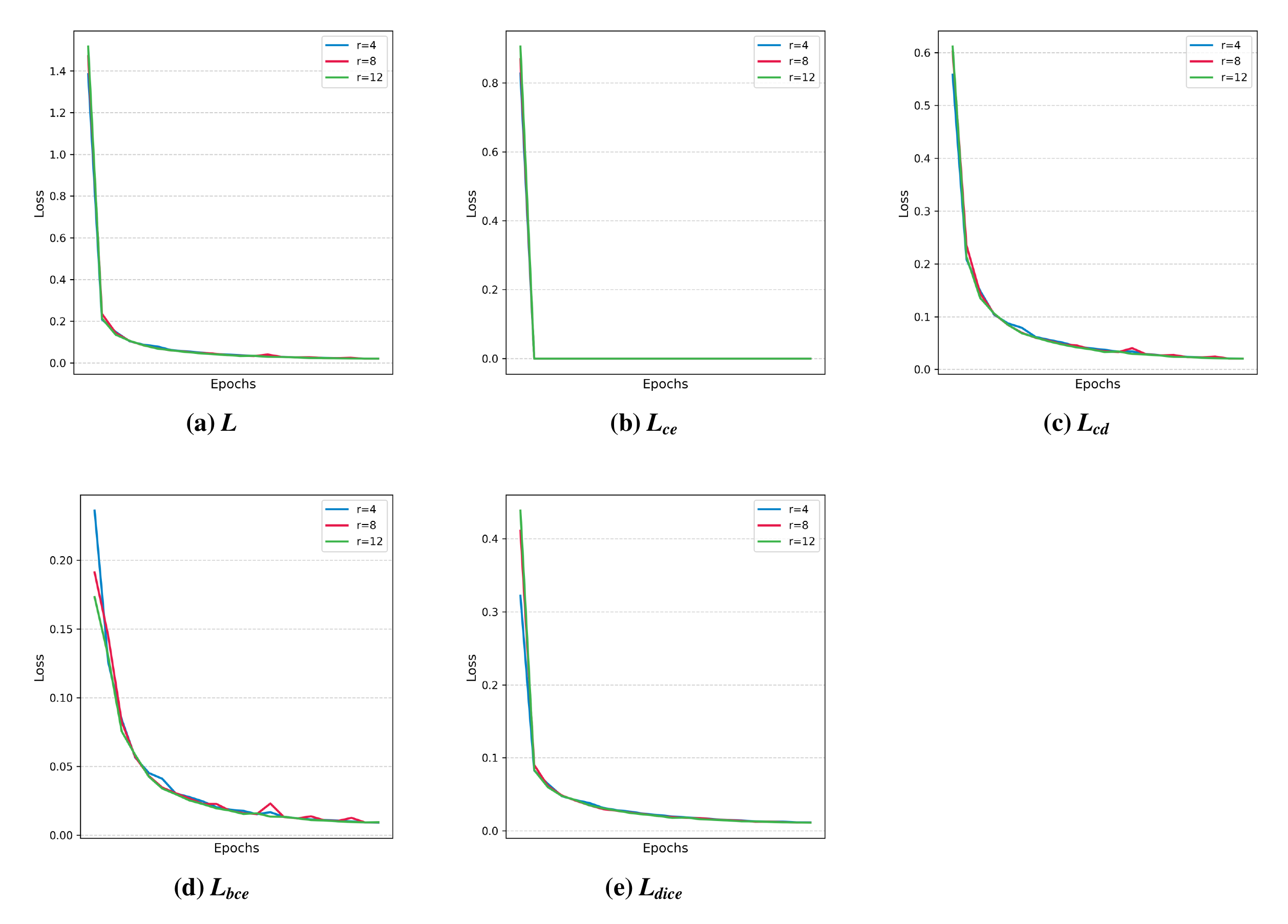}
    \caption{Loss curves of ReasonCD under different LoRA ranks on the BCDD training set.}
    \label{fig13}
\end{figure}

The training loss curves of ReasonCD under different LoRA ranks on the BCDD training set are shown in Figure \ref{fig13}. The blue, red, and green curves represent the loss curves when the rank is 4, 8, and 12.

The experimental results show that the loss curves under different LoRA ranks are almost identical, indicating that the LoRA rank does not significantly impact the loss value. However, there are subtle differences. In Figure \ref{fig13}(d), the loss curve for higher ranks descends more smoothly, with lower frequency of oscillation, aiding in more stable loss reduction.

\begin{figure}%[h]  Figure 14
    \centering
    \includegraphics[width=0.5\textwidth]{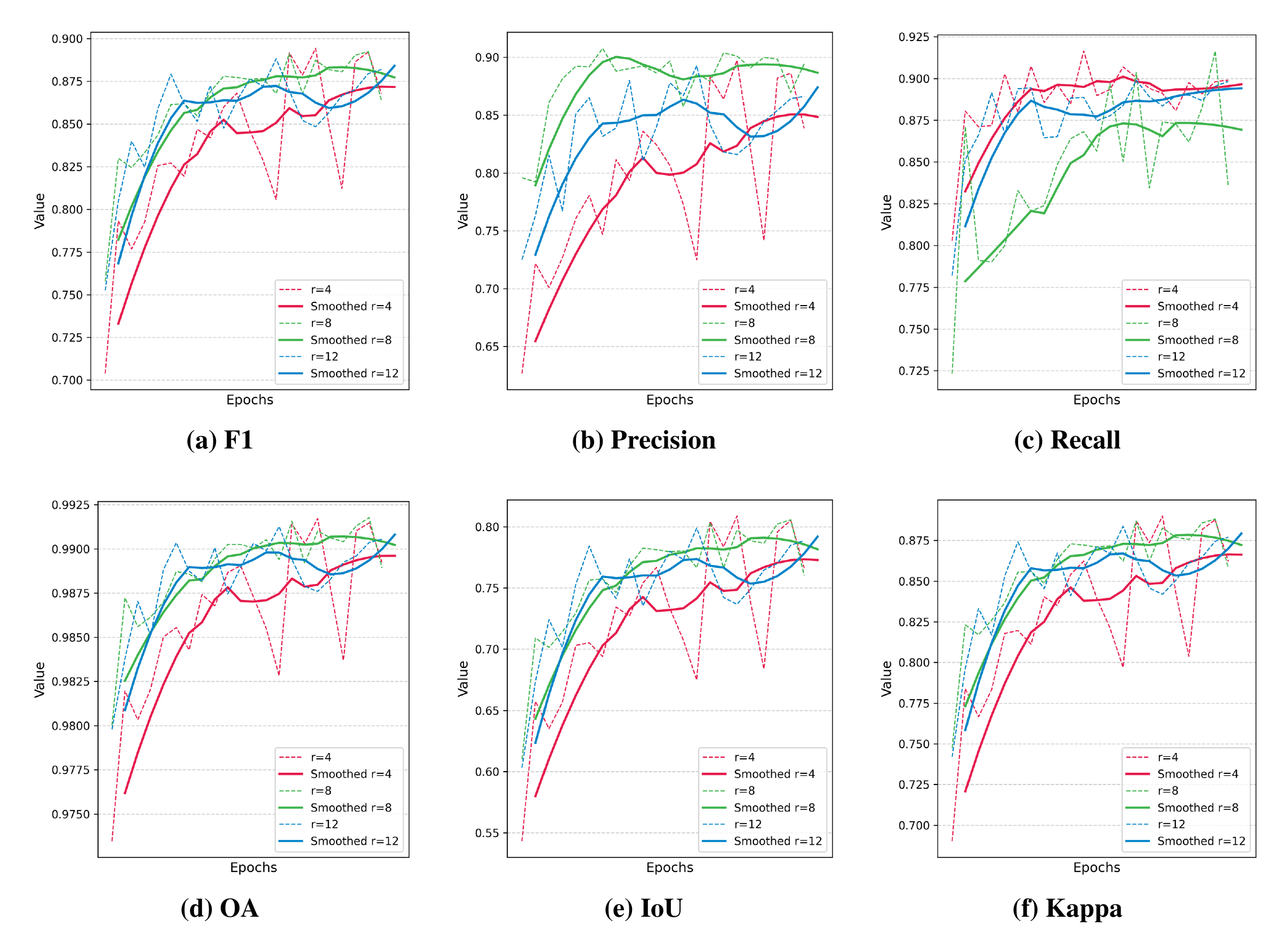}
    \caption{Training metric curves on the BCDD validation set of ReasonCD under different LoRA ranks set.}
    \label{fig14}
\end{figure}

The validation metric curves of ReasonCD under different LoRA ranks on the BCDD validation set are shown in Figure \ref{fig14}. The red, green, and blue curves represent the metric curves when the rank is 4, 8, and 12, respectively, where dashed lines represent the original curves, and solid lines represent the smoothed curves.

The experimental results show that there are also obvious differences in the metric curves under different LoRA ranks. When $r$=8, ReasonCD's metrics overall show better performance, again indicating that when the rank is 8, ReasonCD's learning ability is stronger. It can also be observed that when the rank is lower, the oscillation amplitude of each metric curve is larger, and training stability is relatively poorer compared to higher ranks.

\subsection{ReasonCD Reasoning Change Detection Visualization Experiment}
In the above experiments, this paper has verified the feasibility of ReasonCD for change detection tasks and found relatively optimal model configurations. The ultimate goal of ReasonCD is to infer change maps corresponding to different user-implicit CRoI tasks and, if needed, explain the reasoning process. The purpose of this experiment is to verify whether ReasonCD can achieve the expected goals, promoting the intelligent development of future change detection tasks. Experimental details and results are as follows:

To train ReasonCD's reasoning ability, this paper annotates part of the SECOND dataset with implicit CRoI description texts and their corresponding output texts as the training set. The visualization content for ReasonCD's reasoning results primarily includes three task types: \textbf{Non-reasoning Prediction}, \textbf{Implicit CRoI Reasoning Prediction}, and \textbf{Implicit CRoI Reasoning Prediction with Explanation}. Non-reasoning prediction refers to using simple, explicit CRoI textual descriptions to prompt ReasonCD to detect changes corresponding to that CRoI. The purpose of designing the non-reasoning prediction experiment is to demonstrate that ReasonCD, while possessing reasoning capabilities, still retains the ability for explicit semantic understanding.

ReasonCD's model configuration and training configuration remain consistent with those in Subsection C. Here, instead of taking the model weights at overfitting, the experiment takes the weights from the iteration with the lowest loss after 100 epochs as the model parameters for subsequent reasoning.

\begin{figure*}[h]%[h]  Figure 15
\centering
\includegraphics[width=7in]{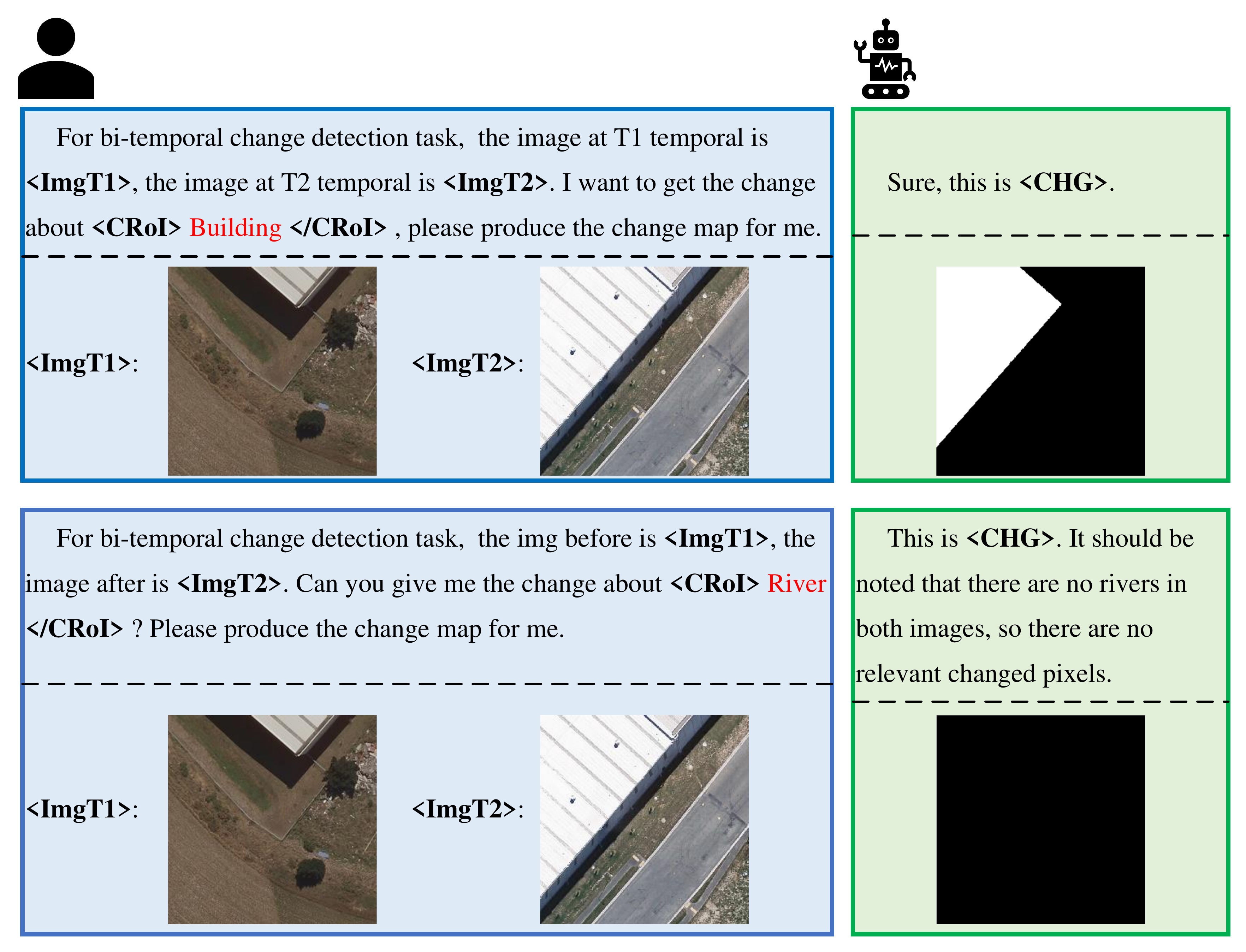}
\caption{Visualization results of ReasonCD performing non-reasoning prediction.}
\label{fig15}
\end{figure*}

The visualization results of ReasonCD performing non-reasoning prediction are shown in Figure \ref{fig15}. The experimental results show that ReasonCD supports change detection tasks using only simple, explicit CRoI textual descriptions as in AUWCD. Furthermore, for CRoIs not present in the image, ReasonCD can also make judgments and provide corresponding prompts.

% \begin{figure}%[h]  Figure 15
%     \centering
%     \includegraphics[width=0.5\textwidth]{fig/fig15.pdf}
%     \caption{Visualization results of ReasonCD performing non-reasoning prediction.}
%     \label{fig15}
% \end{figure}

In this experiment, the selected bitemporal images include changes in buildings, roads, and trees. In the first row of the figure \ref{fig15}, we first use "building" as the CRoI textual description to prompt ReasonCD for building change detection; in the second row, we use "River," which is not present in the image, as the CRoI textual description to prompt ReasonCD for water body change detection.

\begin{figure*}[h]%[h]  Figure 16
\centering
\includegraphics[width=7in]{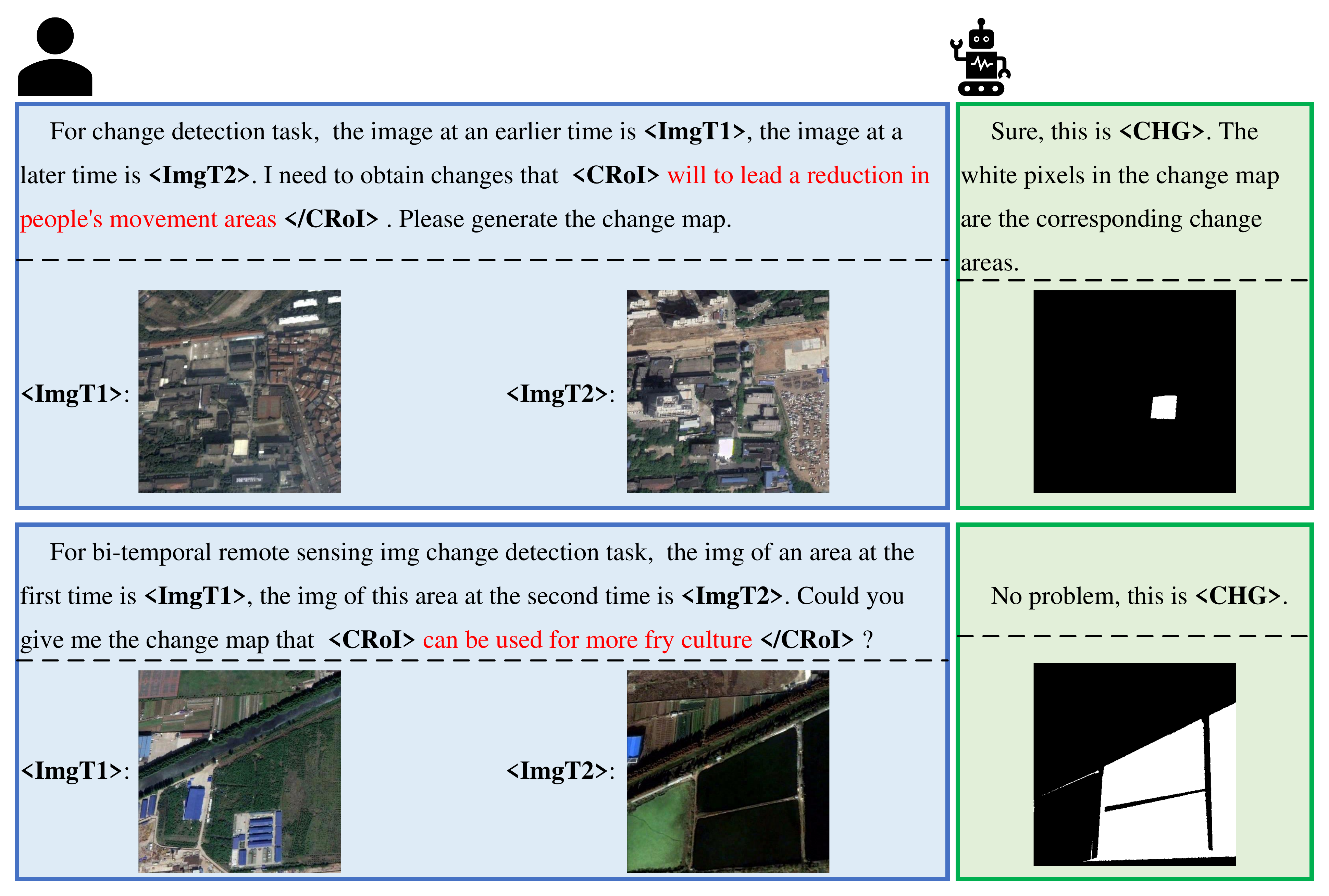}
\caption{Visualization results of ReasonCD performing implicit CRoI reasoning prediction.}
\label{fig16}
\end{figure*}

The visualization results of ReasonCD performing implicit CRoI reasoning prediction are shown in Figure \ref{fig16}. The experimental results show that besides using explicit CRoI textual descriptions, ReasonCD can also excellently complete reasoning for implicit CRoI textual descriptions.
% \begin{figure}%[h]  Figure 16
%     \centering
%     \includegraphics[width=0.5\textwidth]{fig/fig16.pdf}
%     \caption{Visualization results of ReasonCD performing implicit CRoI reasoning prediction.}
%     \label{fig16}
% \end{figure}

\begin{figure*}[h]%[h]  Figure 17
\centering
\includegraphics[width=7in]{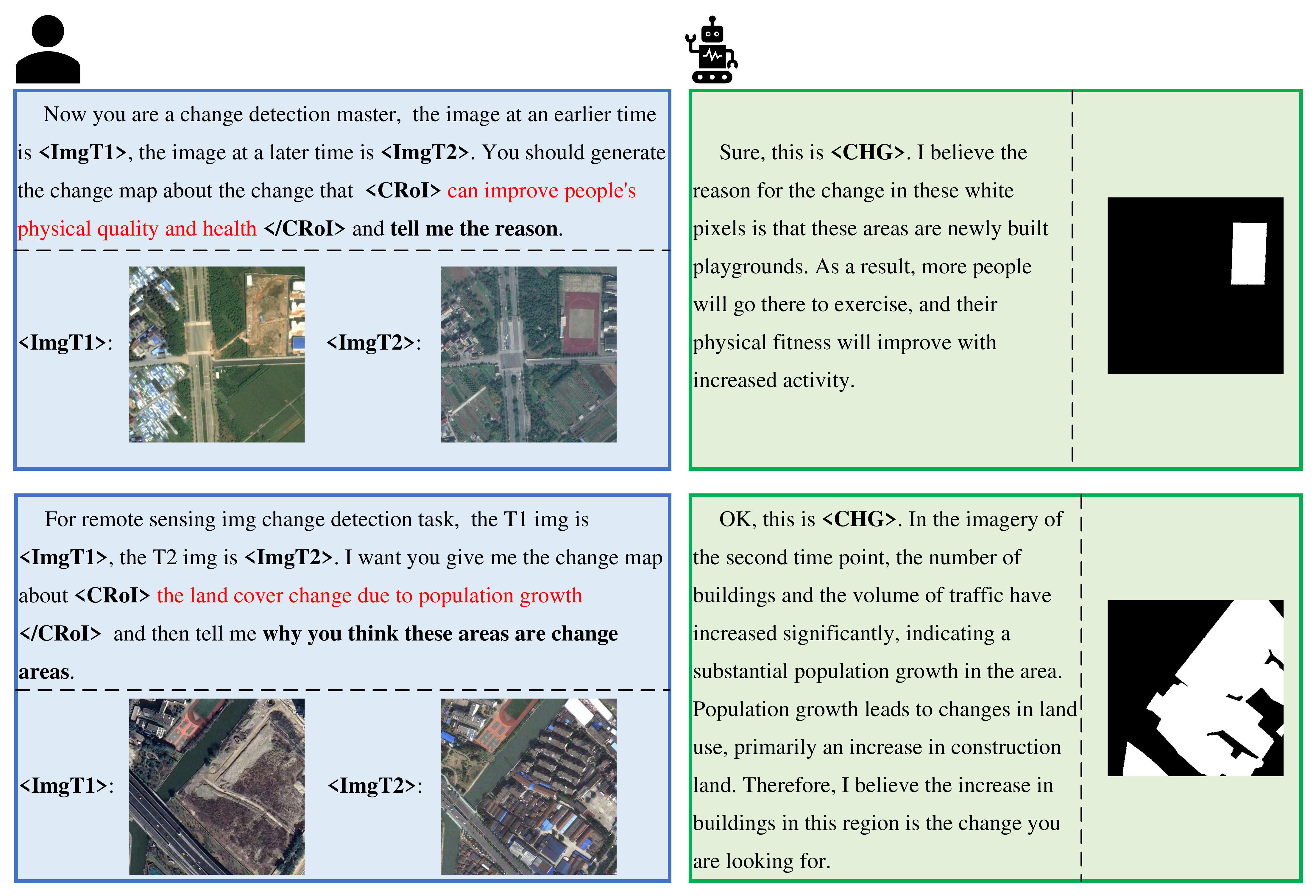}
\caption{Visualization results of ReasonCD performing implicit CRoI reasoning prediction with explanation.}
\label{fig17}
\end{figure*}

The visualization results of ReasonCD performing implicit CRoI reasoning prediction with explanation are shown in Figure \ref{fig17}. The experimental results show that ReasonCD can not only achieve the implicit CRoI reasoning task but also explain its reasoning process, and from the results, the explanations appear reasonable.

Therefore, based on the above three experimental results, ReasonCD not only achieves its design goal of implicit CRoI reasoning but can also explain its reasoning process to assist human decision-making regarding its inference results.

\section{Conclusion and future work}
The performance of current semantics-first paradigm (SeFi-CD) change detection methods, typified by AUWCD, is highly sensitive to the quality of CRoI textual descriptions, requiring simple and precise descriptions to achieve good change detection results. However, in practice, for many tasks, CRoI semantics are often implicit rather than explicit. Using AUWCD in such cases leads to severe degradation in change detection performance. To address this issue and further advance the development of the SeFi-CD paradigm, this paper proposes ReasonCD, a change detection network for mining implicit CRoI semantics. ReasonCD is a multimodal change detection large model based on the large language model LLaMA2. Its core idea is to use LLaMA2 as the reasoning core for implicit semantics to infer high-dimensional CRoI semantic features. The inferred features are then passed through a ChangeMap Decoder to ultimately decode change maps corresponding to the implicit CRoI textual description. For the image features input to the ChangeMap Decoder, this paper designs a Multi-scale Feature Fusion Module $fu(\bullet)$ to compensate for the lack of low-level visual features when using only the SAM image encoder. To validate the performance and feasibility of ReasonCD, this chapter also conducts experiments using the BCDD and SECOND datasets. Experiments prove that the designed Multi-scale Feature Fusion Module $fu(\bullet)$ significantly improves ReasonCD's change detection performance. Furthermore, this chapter annotates part of the SECOND dataset with reasoning training samples to train ReasonCD's reasoning ability. Experiments demonstrate that ReasonCD not only accomplishes the implicit CRoI reasoning task but can also explain its reasoning process to assist human decision-making. Therefore, ReasonCD further advances the intelligence level of SeFi-CD paradigm methods to better adapt to the needs of remote sensing intelligent interpretation in the era of artificial intelligence.

It should be noted that for ReasonCD, this paper only trained its functionalities for non-reasoning prediction, implicit CRoI reasoning prediction, and implicit CRoI reasoning prediction with explanation. In fact, ReasonCD also supports tasks such as change description and change analysis\cite{change_caption1,change_caption2}. If ReasonCD's framework can be utilized in the future to unify all these tasks, and large model technologies like LangChain\cite{langchain} can be employed to analyze and process its output results, constructing a general-purpose change detection task intelligent agent would be a very meaningful research direction. Additionally, the training of ReasonCD in this paper only used optical high-resolution remote sensing imagery and text data, two modalities. Remote sensing data modalities are diverse (e.g., infrared, hyperspectral, SAR, point clouds, etc.). Therefore, unifying various remote sensing data modalities into the ReasonCD framework is also a valuable future research direction. Finally, recent breakthroughs like DeepSeek\cite{deepseekv3,deepseekr1} have shown that powerful reasoning capabilities can be trained using reinforcement learning alone. Thus, proposing a reinforcement learning strategy suitable for enhancing ReasonCD's reasoning ability is equally meaningful future work.

Furthermore, there is currently no large-scale change detection dataset comprehensively designed for the evaluation and training of SeFi-CD paradigm methods. Constructing such a dataset is of great significance for the effective evaluation of future SeFi-CD methods and the training of more powerful models.

%\bibliographystyle{unsrt}
%\bibliography{cit}

\bibliography{ref} 
\bibliographystyle{IEEEtran}

% argument is your BibTeX string definitions and bibliography database(s)
%\bibliography{IEEEabrv, IEEEtran/bibtex/IEEEexample}

\begin{IEEEbiography}[{\includegraphics[width=1in,height=1.25in,clip,keepaspectratio]{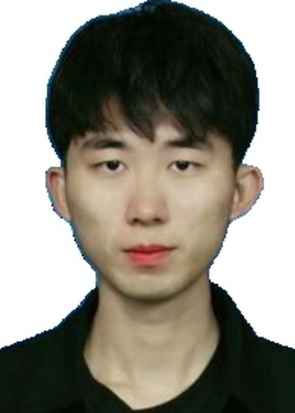}}]{Zhenyang Huang} is currently working at Beijing Institute of Tracking and Telecommunications Technology. He received a master's degree with the School of Geosciences and Info-Physics, Central South University, Changsha, China,in 2025. His research interests include computer vision, deep learning, multimodal large models, and remote sensing image change detection.
\end{IEEEbiography}

\begin{IEEEbiography}[{\includegraphics[width=1in,height=1.25in,clip,keepaspectratio]{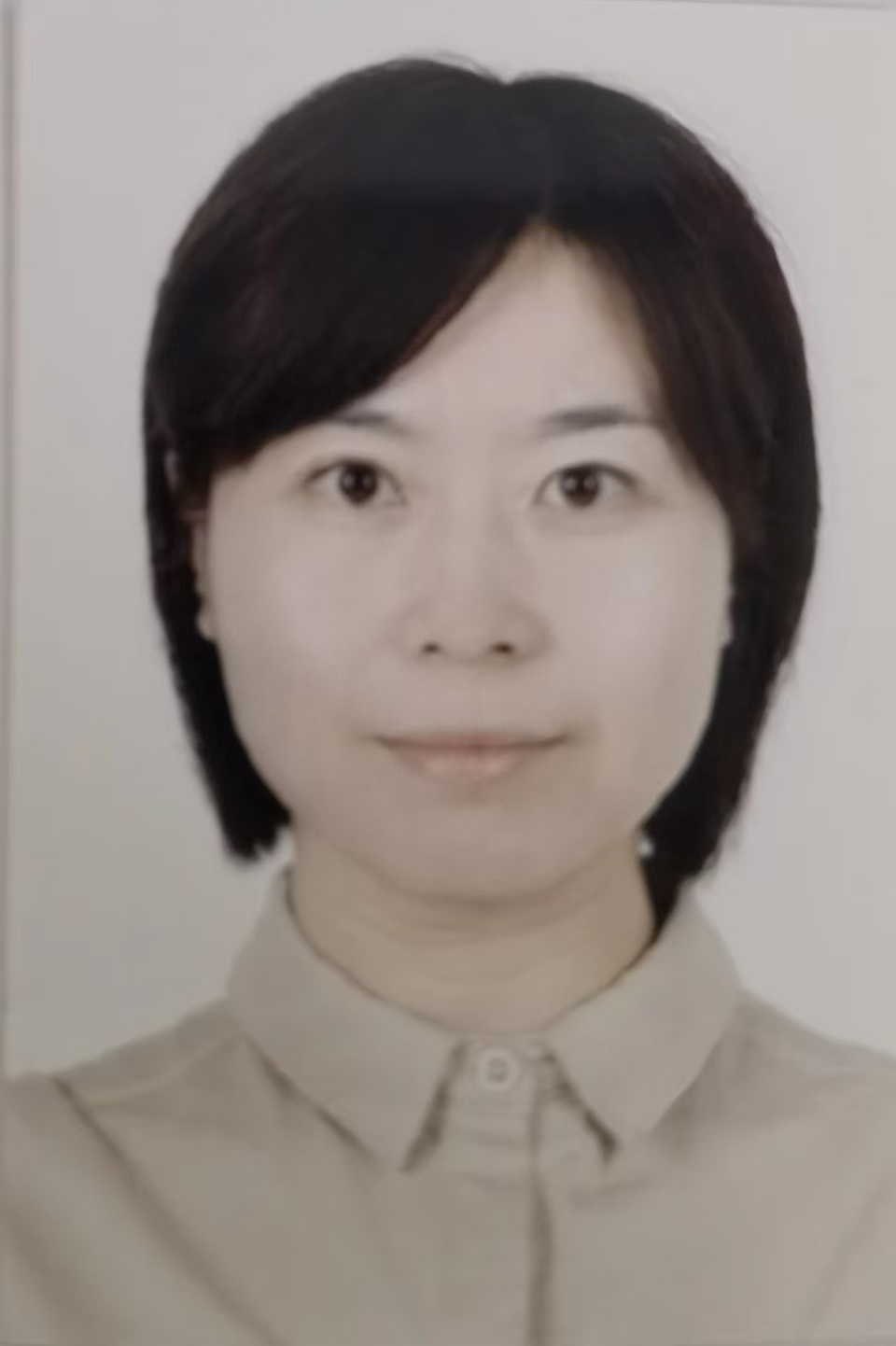}}]{Xiao Yu} received a master's degree from the School of computer Science , National University of Defense Technology, Changsha, China. Her research interest include cloud computing, remote sensing image data processing, and information fusion.
\end{IEEEbiography}

\begin{IEEEbiography}[{\includegraphics[width=1in,height=1.25in,clip,keepaspectratio]{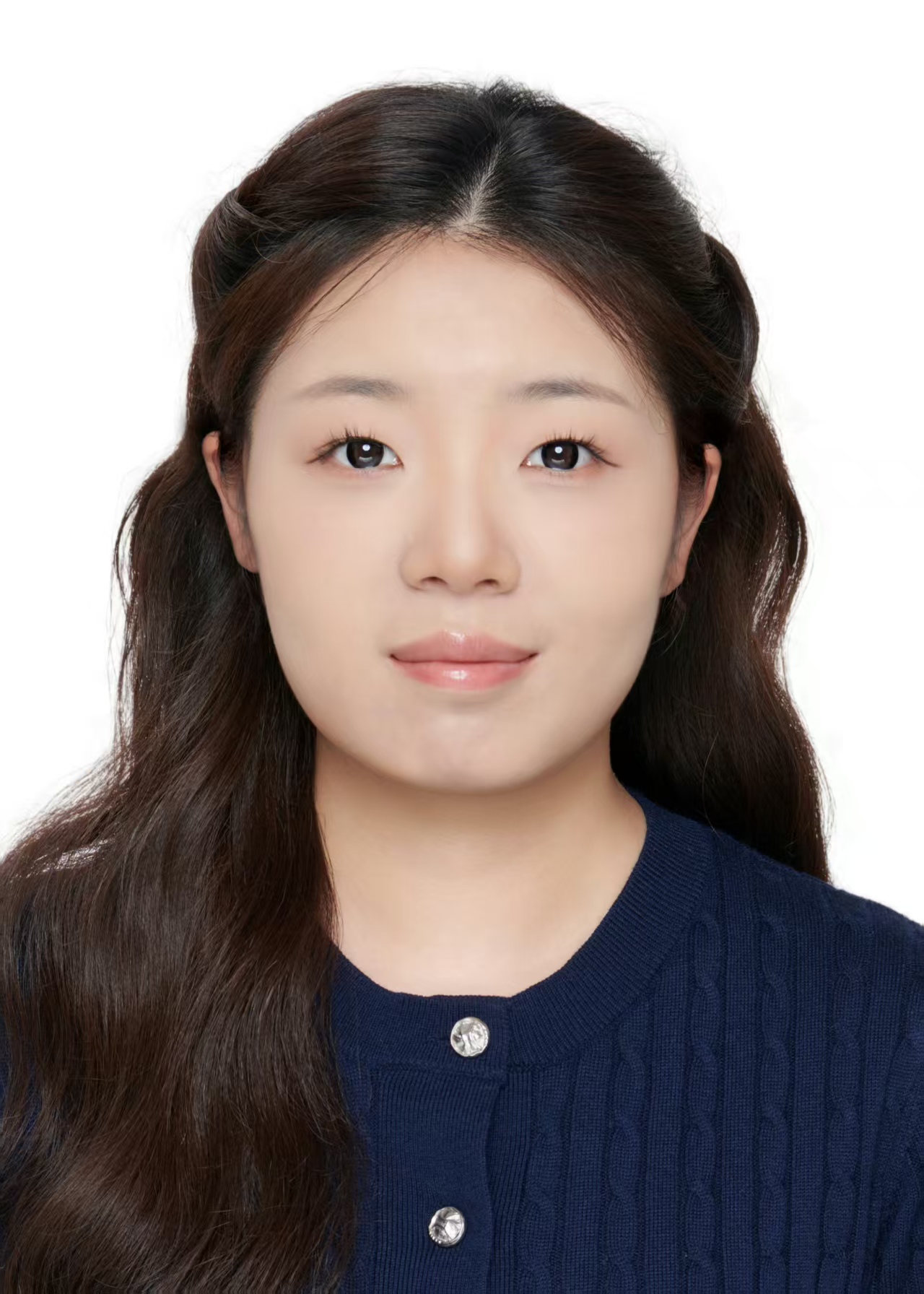}}]{Yi Zhang} is currently working at Beijing Institute of Tracking and Communication Technology.She received a master's degree from the School of Earth Sciences and Engineering at Shandong University of Science and Technology,Qingdao,China,in 2022.Her reserch interests include deep learning, cloud technology and remote sensing data processing.
\end{IEEEbiography}

\begin{IEEEbiography}[{\includegraphics[width=1in,height=1.25in,clip,keepaspectratio]{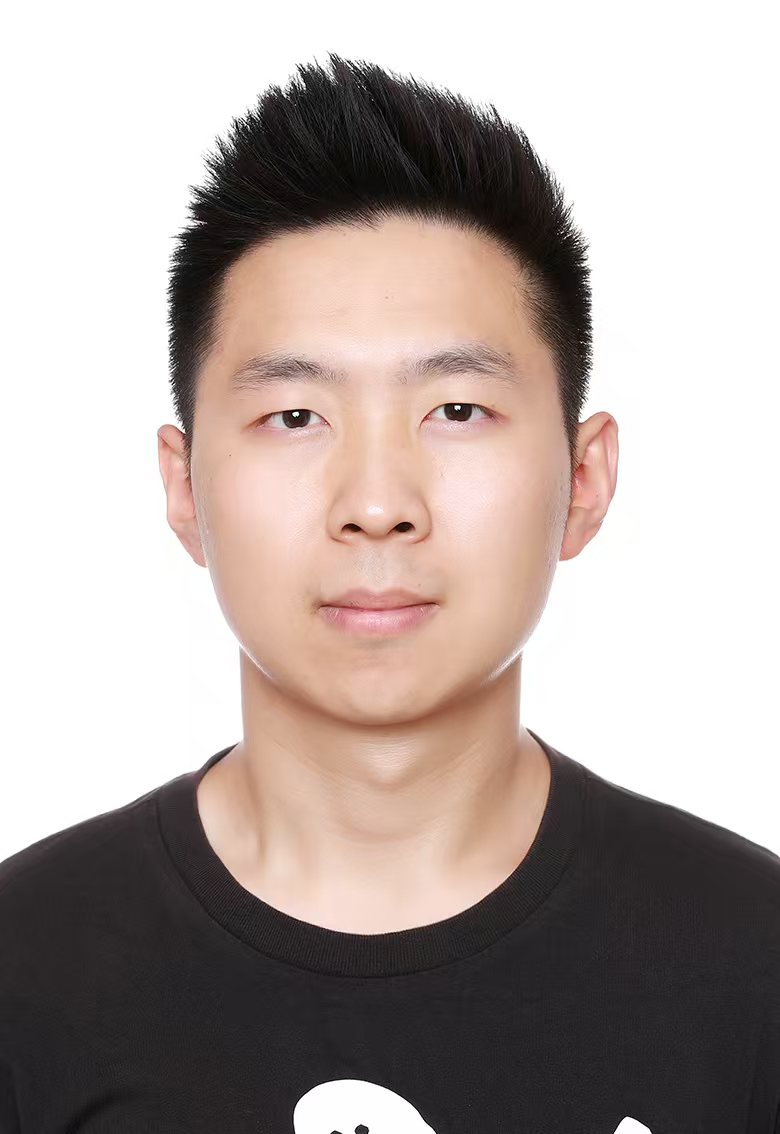}}]{Decheng Wang} is currently working at Beijing Institute of Tracking and Telecommunications Technology. He received the Ph.D. degree with the Space Engineering University, Beijing, China. His research interests include computer vision, deep learning, multimodal large models, and remote sensing image change detection.
\end{IEEEbiography}

\begin{IEEEbiography}[{\includegraphics[width=1in,height=1.25in,clip,keepaspectratio]{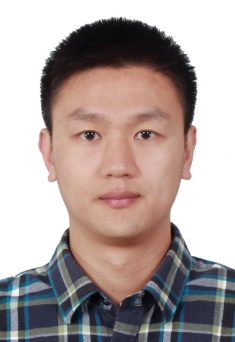}}]{Hang Ruan} is currently an assistant researcher in Beijing Institute of Tracking and Telecommunications Technology, Beijing, China. He received a B.S. degree in Zhejiang University in 2008, and the Ph.D. degree with the Department of Photoelectric Equipment, Academy of Equipment, Beijing, China, in 2013. His main research interests include intelligent interpretation and change detection of remote sensing images.
\end{IEEEbiography}

\end{document}